\documentclass{article}

\usepackage{microtype}
\usepackage{graphicx}
\usepackage{subcaption}
\usepackage{booktabs} 

\usepackage{hyperref}
\usepackage{float}

\usepackage{iftex}
\ifPDFTeX
\else\ifLuaTeX
  \pdfvariable objcompresslevel=0
\fi\fi

\usepackage[preprint]{icml2026}

\usepackage{amsmath}
\usepackage{amssymb}
\usepackage{mathtools}  
\usepackage{amsthm}
\usepackage{xspace}
\usepackage{listings}

\lstdefinelanguage{JavaScript}{
  keywords={break,case,catch,const,continue,debugger,default,delete,do,else,export,finally,for,function,if,import,in,instanceof,let,new,return,super,switch,this,throw,try,typeof,var,void,while,with,yield,async,await,class,extends},
  keywordstyle=\bfseries,
  sensitive=true,
  morecomment=[l]{//},
  morecomment=[s]{/*}{*/},
  morestring=[b]",
  morestring=[b]',
}

\usepackage{colortbl}
\usepackage{tikz} 
\usepackage{multirow}
\usepackage{longtable}
\usepackage{array}
\usepackage{xurl}
\usepackage{enumitem}
\usepackage{tcolorbox}

\usepackage[capitalize,noabbrev]{cleveref}

\theoremstyle{plain}

\theoremstyle{definition}

\theoremstyle{remark}

\usepackage[textsize=tiny]{todonotes}

\icmltitlerunning{See, Plan, Snap: Evaluating Multimodal GUI Agents in Scratch}

\newcommand{\bench}{\textsc{ScratchWorld}\xspace}

\begin{document}

\twocolumn[
  \icmltitle{See, Plan, Snap: Evaluating Multimodal GUI Agents in Scratch}



  \icmlsetsymbol{equal}{*}

  \begin{icmlauthorlist}
    \icmlauthor{Xingyi Zhang}{ecnu}
    \icmlauthor{Yulei Ye}{ecnu}
    \icmlauthor{Kaifeng Huang}{tongjiu}
    \icmlauthor{Wenhao Li}{tongjiu}
    \icmlauthor{Xiangfeng Wang}{ecnu}
  \end{icmlauthorlist}

  \icmlaffiliation{tongjiu}{Tongji University}
  \icmlaffiliation{ecnu}{East China Normal University}

  \icmlcorrespondingauthor{Xingyi Zhang}{xingyi.zhang@stu.ecnu.edu.cn}
  \icmlcorrespondingauthor{Yulei Ye}{52295901055@stu.ecnu.edu.cn}
  \icmlcorrespondingauthor{Kaifeng Huang}{kaifengh@tongji.edu.cn}
  \icmlcorrespondingauthor{Wenhao Li}{whli@tongji.edu.cn}
  \icmlcorrespondingauthor{Xiangfeng Wang}{xfwang@cs.ecnu.edu.cn}

  \icmlkeywords{Machine Learning, ICML}

  \vskip 0.3in
]



\printAffiliationsAndNotice{}  

\begin{abstract}
  Block-based programming environments such as Scratch play a central role in low-code education, yet evaluating the capabilities of AI agents to construct programs through Graphical User Interfaces (GUIs) remains underexplored. We introduce \bench, a benchmark for evaluating multimodal GUI agents on program-by-construction tasks in Scratch. Grounded in the Use-Modify-Create pedagogical framework, \bench comprises 83 curated tasks spanning four distinct problem categories: Create, Debug, Extend, and Compute. To rigorously diagnose the source of agent failures, the benchmark employs two complementary interaction modes: primitive mode requires fine-grained drag-and-drop manipulation to directly assess visuomotor control, while composite mode uses high-level semantic APIs to disentangle program reasoning from GUI execution. To ensure reliable assessment, we propose an execution-based evaluation protocol that validates the functional correctness of the constructed Scratch programs through runtime tests within the browser environment. Extensive experiments across state-of-the-art multimodal language models and GUI agents reveal a substantial reasoning--acting gap, highlighting persistent challenges in fine-grained GUI manipulation despite strong planning capabilities.
\end{abstract}

\section{Introduction}

The movement toward software development democratization has created unprecedented opportunities for low-code paradigms~\citep{gomesLowCodeDevelopmentPlatforms2022}. 
By transforming complex symbolic representations into intuitive visual interfaces, low-code technologies democratize software development, extending it beyond professional developers to a broader population of non-expert users. 
The recent years have seen its widespread adoption in domains such as data analysis~\citep{saLowCodeApproach2024}, scientific simulation~\citep{bucaioniModellingLowcodeDevelopment2022}, and game development~\citep{10.1145/3640310.3674099}. 
Yet the potential of this paradigm extends far beyond industrial applications; it has also opened new avenues in education. 
\textit{Scratch}~\citep{resnickScratchProgrammingAll2009}, a programming language designed specifically for children and adolescents, has emerged as a representative and influential system. 
It features block-based visual interaction interfaces and a drag-and-drop programming paradigm designed to support playful, game-like learning experiences while fostering creativity and computational thinking among children.

Younger learners often struggle with 
(i) forming a stable understanding of program execution and debugging~\citep{e94d570b621749cb96f41c81e9a50588, 10.1111/j.1467-8624.2010.01499.x}, and 
(ii) accurate interaction with the Graphical User Interface (GUI) (e.g., navigation and drag-and-drop based block manipulation)~\citep{digital4010002, DONKER2007257} in Scratch.
Therefore, this necessitates support for younger learners from experienced instructors. 
As teachers are often constrained by scheduled classes and limited time per student, an AI-assisted teacher appears to be a more favorable option. 

To address this resource constraint, several Large Language Model (LLM)-based assistants~\citep{siStitchStepbystepLLM2025, feinLitterBoxExtensibleFramework2025, drugaScratchCopilotSupporting2025b} for Scratch have been proposed. 
They can provide textual debugging and modification suggestions, or even directly manipulate the underlying code representation to update the block structures. 
However, these systems create a split-attention demand~\citep{Ayres2012}: 
learners must mentally integrate text-based guidance with spatially located GUI actions (e.g., click, drag-and-drop), which adds extraneous cognitive load unrelated to the programming goal~\citep{069d6b7c-5317-393c-a548-0192069176ab}.

Fortunately, recent advances in GUI agents~\citep{agasheAgentOpenAgentic2024b, qinUITARSPioneeringAutomated2025a} offer a promising alternative.
These agents can perceive graphical interfaces through screenshots, interpret visual and textual elements, reason about task solutions through multi-step decomposition, and automatically execute GUI operations such as clicking, typing, and drag-and-drop.
Unlike chat-based assistants that offer documented solutions, GUI agents directly demonstrate how to do it through observable interface interactions, eliminating the cognitive overhead of translating verbal instructions into spatial actions.
For Scratch learners, this hands-on guidance can accelerate procedural proficiency in block manipulation while freeing cognitive resources for higher-order computational thinking.

While GUI agents offer this potential, systematic evaluation on block-based programming remains underexplored.
Existing GUI agent benchmarks for web~\citep{zhouWebArenaRealisticWeb2024d, dengMind2WebGeneralistAgent2023a}, desktop~\citep{nayak2025uivisiondesktopcentricguibenchmark,kapoorOmniACTDatasetBenchmark2025a, xieOSWorldBenchmarkingMultimodal2024a}, and mobile~\citep{liEffectsDataScale2024, rawlesAndroidWorldDynamicBenchmarking2025} platforms primarily focus on navigational or retrieval-oriented tasks such as web browsing, file management, and app navigation.
In contrast, Scratch demands \textit{program-by-construction}, where agents must assemble executable programs through long-horizon, compositional drag-and-drop operations that require not only precise visual grounding but also structured reasoning about program logic, resulting in a clear benchmarking gap.

To address this gap, we introduce \bench, the first benchmark for evaluating GUI agents on program-by-construction in block-based environments.
\bench comprises 83 carefully curated tasks spanning four problem categories: \textit{Create}, \textit{Debug}, \textit{Extend}, and \textit{Compute}, which align with the Use-Modify-Create pedagogical framework~\citep{lytleUseModifyCreate2019}.
To rigorously diagnose whether failures stem from logical reasoning deficits or visuomotor execution errors, \bench employs a dual-mode evaluation protocol.
In primitive mode, agents receive screenshots with indexed element lists and must execute low-level UI primitives (click, drag-and-drop, type).
In composite mode, agents receive pseudocode representations and execute high-level semantic commands (add block, connect blocks), isolating program logic from GUI manipulation.
All tasks are validated through execution-based runtime tests to ensure functional correctness (Section \ref{sec:evaluation}).

\begin{figure}[htb!]
    \centering
    \includegraphics[width=.85\linewidth]{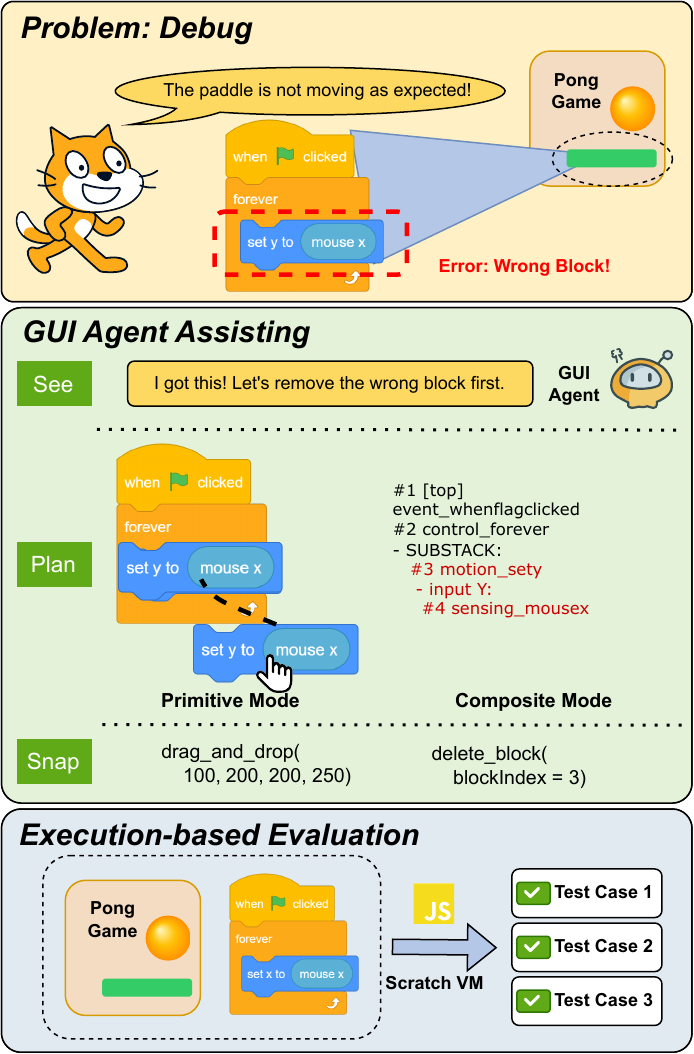}
    \caption{Overview of \bench evaluation workflow through a Debug task: fixing incorrect paddle control in Pong. Agents interact via two modes: primitive mode using GUI operations (drag-and-drop) or composite mode using high-level APIs (delete block). Execution-based evaluation leverages Scratch VM to validate functional correctness.}
\end{figure}

Through initial experiments, we observe a substantial performance gap between the two modes: while state-of-the-art (SOTA) models achieve over 78\% success rate in composite mode, their performance drops to merely 14\% in primitive mode.
To explain the low primitive-mode success, we first isolate drag-and-drop precision via a Single-Step Drag Benchmark, revealing that even single-step actions frequently fail. Given this finding, we then investigate whether perception deficits account for these failures via a Visual Perception QA Benchmark.

Through extensive empirical evaluation on \bench, we highlight several phenomena:

\noindent\textbf{Reasoning--Acting gap:} We observe a severe dichotomy in agent capabilities. The best result in composite mode reaches 78.31\% overall success rate (Claude-Sonnet-4.5), while the best result in primitive mode drops to 14.46\% (AWM + Claude-Sonnet-4.5). This quantifies a critical bottleneck: models can plan the program logic but fail to reliably manipulate the interface.

\noindent\textbf{Endpoint Localization Bottleneck:} To explain this gap, results of Single-Step Drag Benchmark show that even when the start point is given, endpoint localization remains poor. With labeled start positions, end-component accuracy stays low (about 23--32\%), indicating that fine-grained spatial grounding for ``where to drop'' is the dominant failure mode.

\noindent\textbf{Perception is Not Sufficient:} High accuracy on static Visual Perception QA Benchmark (up to 90.5\%) does not explain drag errors in dynamic GUI manipulation. This shows that static perception alone is insufficient; precise closed-loop execution remains the key challenge in Scratch drag-and-drop interactions.

In short, \bench separates ``can you figure out the right program?'' from ``can you actually build it in the GUI?'' and our experiments show the latter is the real blocker today. 
Even when models solve the logic in composite mode (\(>78\%\) SR), they largely collapse under primitive, pixel-level interaction (\(\sim14\%\) SR), revealing a pronounced reasoning--acting gap. 
Diagnostics further point to endpoint localization (``where to drop'') as the dominant failure mode, and strong static perception (up to 90.5\% QA accuracy) still does not translate into reliable closed-loop drag-and-drop execution. 
Together, these findings argue that progress on Scratch-style program-by-construction will hinge less on better planning and more on robust, snap-aware, high-precision interaction policies.

\section{Related Work}

\noindent\textbf{Intelligent Assistance in Scratch.}
Recent works have begun to explore LLM-based intelligent assistance for real-time and personalized support in Scratch. 
\citet{drugaScratchCopilotSupporting2025b} integrated LLMs into Scratch as a chatbot to offer suggestions on project ideas and manipulation strategies; \citet{feinLitterBoxExtensibleFramework2025} used static analysis to detect common bugs and code smells in Scratch project and used LLMs to generate explanation and applicable fixes; \citet{siStitchStepbystepLLM2025} analyzed difference between current project and golden project to provide step-by-step instructions. 

In contrast to intelligent assistance in the form of chatbot, a more straightforward assistance is to receive direct hands-on GUI operation. 
GUI agents can help, which is demonstrated to be effective in web navigation~\citep{zhouWebArenaRealisticWeb2024d} and smartphone control~\citep{rawlesAndroidWorldDynamicBenchmarking2025} tasks, other than Scratch.
Our benchmark evaluates GUI agents that can autonomously perform click and drag-and-drop operations, providing a more direct solution to assistance on foundational procedural skill of block manipulation itself, which is essential for effective Scratch programming.

\noindent\textbf{Benchmarking GUI Agents.}
Existing GUI agents benchmarks span a variety of platforms: web, desktop and mobile, each focusing on different interaction paradigms and abilities.
\textbf{Web}-based benchmarks~\citep{zhouWebArenaRealisticWeb2024d, dengMind2WebGeneralistAgent2023a, yaoWebShopScalableRealWorld2022} predominantly test browser-based navigation and information manipulation: DOM-targeted clicking, link following, search, pagination, form filling, and state changes. 
\textbf{Desktop} environment is more complex as it contains a wide range of applications and tasks that require complex manipulation skills. 
Existing desktop benchmarks~\citep{nayak2025uivisiondesktopcentricguibenchmark, kapoorOmniACTDatasetBenchmark2025a,xieOSWorldBenchmarkingMultimodal2024a, bonattiWindowsAgentArena2024a} have covered many daily and professional scenarios, thus providing a broad evaluation of an agent's capabilities in desktop environments.
\textbf{Mobile} benchmarks~\citep{liEffectsDataScale2024, rawlesAndroidWorldDynamicBenchmarking2025} are characterized by touch-based interactions and primarily evaluate agents on common smartphone tasks such as app launching, menu navigation, and list scrolling. 

Across categories, existing GUI benchmarks provide valuable coverage of navigation, search, and form-based manipulation, but they leave a critical gap: 
they do not evaluate an agent's ability to perform program-by-construction in a block-based or low-code interface, where precise drag-and-drop composition and structural reasoning over program logic are essential. 
\bench is the first benchmark designed to directly address this gap.

\noindent\textbf{LLM-based Coding Agents.}
Coding LLMs has rapidly emerged as one of the most transformative forces in modern software engineering.
This progress is evident across key tasks, most notably in code generation and program repair, where LLMs have consistently pushed the SOTA. 

In code generation, the field has been defined by influential benchmarks such as MBPP~\citep{austinProgramSynthesisLarge2021b}, HumanEval~\citep{chen2021evaluatinglargelanguagemodels}, and HumanEval-X~\citep{zhengCodeGeeXPreTrainedModel2024}. 
These benchmarks have, in turn, fueled the development of powerful foundation models~\citep{deepseek-aiDeepSeekCoderV2BreakingBarrier2024, huiQwen25CoderTechnicalReport2024, roziereCodeLlamaOpen2024}, sophisticated agentic frameworks~\citep{rasheedCodePoriLargeScaleSystem2024, zhangCodeAgentEnhancingCode2024, wangOpenHandsOpenPlatform2025} and commercial AI-powered coding tools~\citep{githubcopilot, cursor, claudecode}.
In parallel, the field of program repair has seen significant advances, benchmarked by tasks like Defects4J~\citep{just2014defects4j}, QuixBugs~\citep{lin2017quixbugs}, and the large-scale SWE-bench~\citep{jimenez2024swebench}. 
This has spurred progress on specialized agents and models capable of understanding and patching complex codebases~\citep{liPatchPilotCostEfficientSoftware2025, olaussonSelfRepairSilverBullet2024, yangSWEagentAgentComputerInterfaces2024}.

However, these works, spanning both code generation and program repair, share a critical and universal limitation: 
their focus is exclusively confined to text-based programming. 
As a result, a standardized benchmark for block-based programming like Scratch remains notably absent.

\section{ScratchWorld Benchmark}

This section details \bench\footnote{Code and Dataset are available at \url{https://github.com/astarforbae/ScratchWorld}.}, including the evaluation tasks (Section~\ref{sec:eval_tasks}), benchmark construction pipeline (Section~\ref{sec:construction}), interaction modes (Section~\ref{sec:modes}), and evaluation protocol (Section~\ref{sec:evaluation}).

\subsection{Evaluation Tasks}
\label{sec:eval_tasks}

\begin{figure*}[t]
  \centering
  \includegraphics[width=\linewidth]{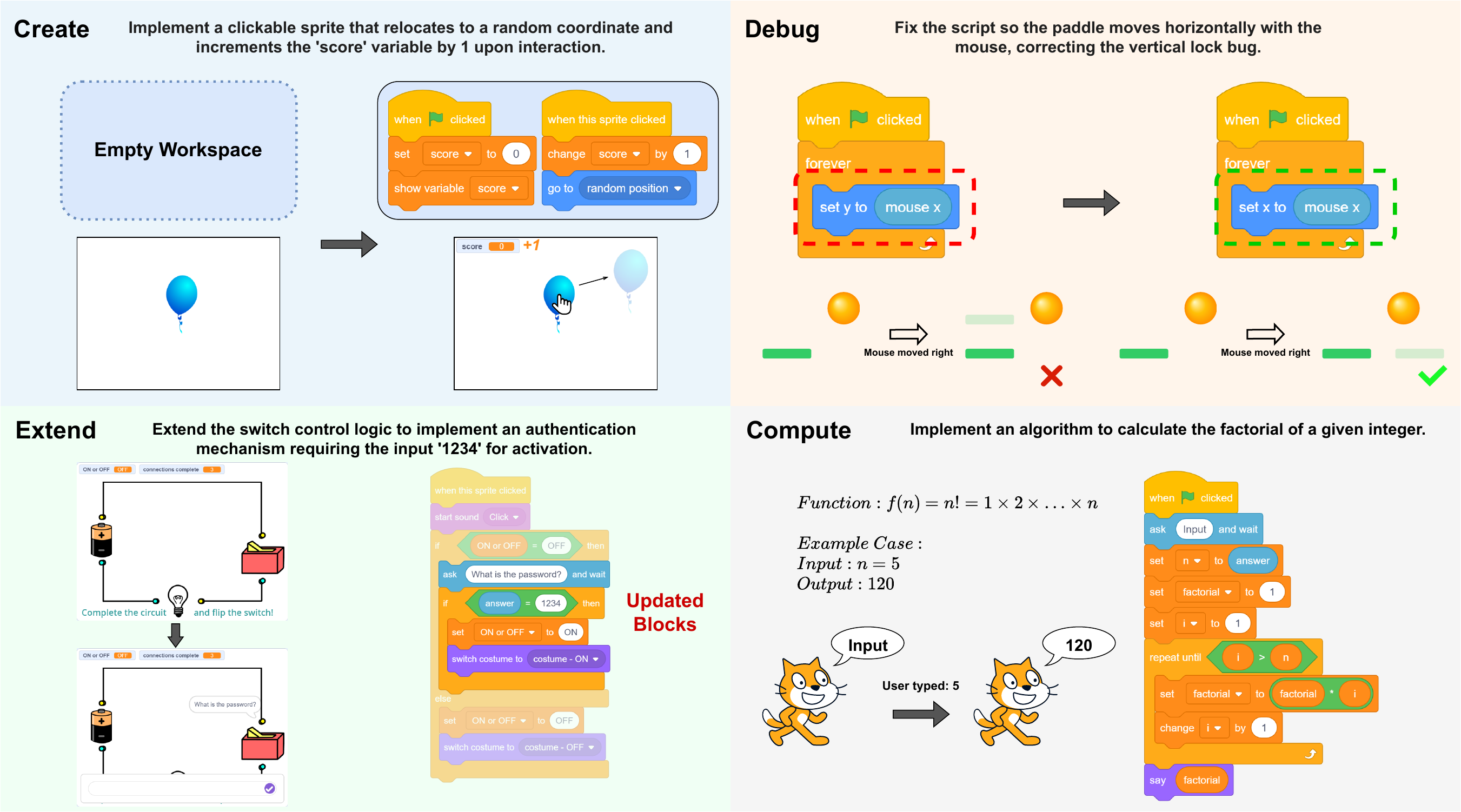}
  \caption{Example tasks for four problem categories in \bench. \textbf{Create} tasks require creating interactive projects from scratch, such as a balloon game. \textbf{Debug} tasks involve diagnosing and correcting bugs, illustrated by fixing a coordinate error in a paddle controller. \textbf{Extend} tasks challenge agents to extend existing functionality, e.g., adding password authentication for a circuit switch. \textbf{Compute} tasks focus on pure computational logic, such as implementing a factorial calculator.}
  \label{fig:tasks}
\end{figure*}

To rigorously evaluate diverse competencies, \bench organizes its 83 tasks into four problem categories, as illustrated in Figure~\ref{fig:tasks}: \textbf{Create}, \textbf{Debug}, \textbf{Extend}, and \textbf{Compute}. These categories align with the Use-Modify-Create pedagogical scaffold~\citep{lytleUseModifyCreate2019}: \textbf{Debug} and \textbf{Extend} tasks both correspond to Modify (repairing buggy code and augmenting functional code with new features, respectively), while \textbf{Create} and \textbf{Compute} tasks map to Create (authoring original projects from scratch, with varying emphasis on interactivity versus algorithmic logic).

Concretely, \textbf{Create} tasks require the agent to synthesize interactive projects with specific visual behaviors from an empty canvas based on natural language descriptions. 
This category rigorously tests high-level planning and compositional reasoning.
\textbf{Debug} tasks present agents with projects containing intentionally injected bugs. 
The agent must diagnose and rectify logic errors in the provided scripts to restore the expected functionality.
\textbf{Extend} tasks challenge agents to extend functional codebases with new functionality based on updated requirements. 
This tests the ability to augment capabilities while preserving existing behavior.
\textbf{Compute} tasks focus on implementing abstract logic and mathematical reasoning using block primitives, prioritizing algorithmic correctness over interactive behaviors.

\subsection{Benchmark Construction Pipeline}
\label{sec:construction}

Each task in \bench comprises four core components: 
\textit{task instruction}, \textit{initial project} (starting state), \textit{golden project} (canonical solution), and a specialized \textit{evaluation script}. 
To ensure high quality and diversity, we developed a semi-automated pipeline comprising three stages:

\textbf{Step 1: Sourcing.} We establish the seed dataset through a hybrid approach. 
For \textbf{Create} and \textbf{Compute} tasks, human experts manually designed a set of seed tasks to define the rigorous structure of instructions and solutions. 
For \textbf{Debug} and \textbf{Extend} tasks, we curated 8 representative projects from the \textit{official Scratch Starter Projects}~\citep{scratchstarterprojects} (see Appendix~\ref{sec:source_projects} for the full list) to serve as high-quality, canonical base codebases.

\textbf{Step 2: Expansion.} We leverage a collaborative multi-model pipeline to scale these seeds. 
We utilize \textbf{DeepSeek-V3}~\citep{deepseekai2025deepseekv3technicalreport} for generating high-level task concepts for \textbf{Create} and \textbf{Compute} categories, brainstorming diverse project requirements.
For \textbf{Debug} and \textbf{Extend} tasks, we serialize Scratch projects into structured text representations (Appendix~\ref{sec:serialization_example}) and employ \textbf{GitHub Copilot}~\citep{githubcopilot}, which analyzes the context of selected representative projects to propose bug injection strategies or feature extensions.
Following human verification of these ideas, we further utilize Copilot to generate \textit{event-driven unit tests}. Based on the finalized task descriptions, Copilot synthesizes JavaScript evaluation scripts that cover all functional requirements specified in the instructions (see Appendix~\ref{sec:appendix_prompts} for part of prompts used in this step).

\textbf{Step 3: Verification.} A strict human-in-the-loop process guarantees validity. Expert annotators review all generated instructions and implement the corresponding \textit{golden projects} to ensure correctness. Crucially, every evaluation script undergoes adversarial testing: it must pass the golden project (true positive) and correctly fail on corrupted solutions (true negatives). Specifically, for \textbf{Debug} tasks, we utilize the buggy \textit{initial project} as the negative case; for other categories, we generate negative cases by removing blocks targeted by the unit tests to ensure the script validates critical logic.

\subsection{Interaction Modes}
\label{sec:modes}

\bench employs two complementary modes to disentangle logical reasoning from visuomotor control.

\noindent\textbf{Primitive mode} is designed to assess an agent's fine-grained GUI manipulation and visual grounding capabilities. We formulate the observation space $\mathcal{O}_{\text{prim}}$ as a \textit{hybrid representation} that augments visual screenshots with textual element lists to mitigate pixel-level ambiguity and align with prior work (e.g., OSWorld~\citep{xieOSWorldBenchmarkingMultimodal2024a}):
\begin{equation}
    \mathcal{O}_{\text{prim}} = \langle \mathbf{I}, \mathcal{E} \rangle,
\end{equation}
where $\mathbf{I} \in \mathbb{R}^{H \times W \times 3}$ represents the full-resolution screenshot, with $H$ and $W$ denoting the height and width respectively, and $3$ representing the RGB color channels.
$\mathcal{E} = \{ (\tau_i, c_i, b_i) \}_{i=0}^{N-1}$ is an indexed list of UI elements, where each element at index $i$ consists of: a \textbf{categorical type} $\tau_i$ (e.g., \texttt{inputs}, \texttt{blocks}); a \textbf{text string} $c_i$ describing the content (e.g., block labels, values); and a \textbf{spatial tuple} $b_i \in \mathbb{N}^4$ specifying the bounding box $(x, y, w, h)$. The index $i$ serves as a unique identifier for interaction targets (see Appendix~\ref{sec:element_list_example} for an example).

Aligning with standard GUI benchmarks~\citep{nayak2025uivisiondesktopcentricguibenchmark, xieOSWorldBenchmarkingMultimodal2024a}, the action space $\mathcal{A}_{\text{prim}}$ consists of atomic UI primitives that mirror human operations:
\begin{equation}
    \mathcal{A}_{\text{prim}} = \mathcal{P}_{\text{mouse}} \cup \mathcal{P}_{\text{keyboard}} \cup \{ \texttt{done}, \texttt{fail} \}.
\end{equation}
Here, $\mathcal{P}_{\text{mouse}}$ includes spatial actions such as \texttt{click}, \texttt{double-click}, \texttt{drag-and-drop}, and \texttt{scroll}; 
$\mathcal{P}_{\text{keyboard}}$ covers \texttt{type}, \texttt{key}, and \texttt{hotkey}.
See Appendix~\ref{sec:primitive_action_space} for complete action specifications.

\noindent\textbf{Composite mode} abstracts away low-level execution to focus on logic planning. Agents receive a structured observation (see Appendix~\ref{sec:comp_observations} for an example) containing variable scopes, targets, and nested block pseudocode. Interaction is performed via high-level \textbf{semantic APIs}, such as \texttt{add\_block}, \texttt{connect\_blocks}, and \texttt{set\_block\_field}, that encapsulate complex UI operations. This design disentangles reasoning from GUI execution, enabling rigorous evaluation of control-flow design and data-flow correctness while eliminating the confound of visuomotor errors. (See Appendix~\ref{sec:composite_action_space} for complete action specifications).

\subsection{Evaluation Protocol}
\label{sec:evaluation}

\paragraph{Execution-based Evaluation Mechanism.}
Following recent advanced benchmarks~\citep{xieOSWorldBenchmarkingMultimodal2024a, bonattiWindowsAgentArena2024a} that utilize functional verification, \bench adopts an execution-based evaluation protocol implemented via Playwright~\citep{playwright}. Unlike general OS tasks where success is often determined by file system changes or terminal outputs, Scratch necessitates rigorous verification of \textit{dynamic internal states} (e.g., sprite coordinates, variable updates during runtime). We interface with the Scratch VM~\citep{scratchvm} exposed in the browser and evaluate each task using a pre-authored JavaScript evaluation script that runs a suite of unit tests (see Appendix~\ref{sec:eval_script_example}). Scripts are generated by the LLM and refined by human experts to cover critical failure modes, ensuring the constructed program meets precise functional requirements beyond visual correctness.

\paragraph{Metrics.}
We define $2$ key metrics to quantify agent performance based on the unit tests.
\textbf{Success Rate (SR)} is a strict metric measuring the percentage of tasks where the agent passes \textit{all} unit tests. 
Let $N$ be the total tasks and $u_i$ be the result of the $i$-th task ($u_i=1$ if all tests pass, else $0$).
\begin{equation}
    SR = \frac{1}{N} \sum_{i=1}^{N} u_i \times 100\%.
\end{equation}
\textbf{Partial Success Rate (PSR)} is a fine-grained metric reflecting how close the agent came to a solution. It is the average ratio of passed unit tests per task. Let $m_i$ be the number of passed tests and $M_i$ be the total tests for task $i$.
\begin{equation}
    PSR = \frac{1}{N} \sum_{i=1}^{N} \frac{m_i}{M_i} \times 100\%.
\end{equation}
While SR indicates perfect completion, PSR provides diagnostic value by revealing partial progress in such long-horizon tasks.

\paragraph{Statistics and Complexity.}
Although \bench comprises a compact set of $83$ tasks, it is designed for high density and complexity.
\textbf{Long-Horizon Reasoning:} Tasks require an average modification of $9.14$ blocks, with the most demanding task requiring the addition of $32$ new blocks. \textbf{Compute} tasks are particularly complex, requiring the construction of $16.96$ functional blocks on average from a minimal scaffold.
\textbf{High Diversity:} The benchmark covers $82$ distinct block types across $8$ categories (Motion, Control, Sensing, etc.), ensuring broad coverage of Scratch's programming primitives. This complexity profile ensures that despite the dataset size, \bench provides a challenging and discriminative testbed for GUI agents.

\begin{table*}[t]
\centering
\small 
\caption{
    Main results on \textsc{ScratchWorld}\xspace across four task categories and two interaction modes. 
    \textbf{Composite Mode} provides high-level semantic APIs for block manipulation, while \textbf{Primitive Mode} requires low-level GUI operations (click, drag-and-drop).
    For each task category, we report Success Rate (\%). 
    Overall performance is measured by \textbf{SR} (Success Rate: percentage of tasks with all tests passed) and \textbf{PSR} (Partial Success Rate: average ratio of passed tests per task).
}
\label{tab:results}
\setlength{\tabcolsep}{8pt} 

\def\heatmaxmix{55} 
\def\heatboxw{7.8mm} 
\newcommand{\heatcell}[2]{%
  \begingroup
  \pgfmathsetmacro{\mix}{abs(#1-50)/50*\heatmaxmix}
  \ifdim #1pt<50pt
    \def\heatcol{red!\mix!white}%
  \else
    \def\heatcol{green!\mix!white}%
  \fi
  \tikz[baseline=(X.base)]\node[
    fill=\heatcol,
    rounded corners=1.2pt,
    inner xsep=0pt,
    inner ysep=0.8pt,
    minimum width=\heatboxw,
    text width=\heatboxw,
    align=center
  ](X){#2};%
  \endgroup
}

\begin{tabular}{lcccccc}
\toprule
& \multicolumn{4}{c}{\textbf{Task Performance (Success Rate \%)}} & \multicolumn{2}{c}{\textbf{Overall Metrics}} \\
\cmidrule(lr){2-5} \cmidrule(lr){6-7}
\textbf{Agent} & \textbf{Create (20)} & \textbf{Debug (20)} & \textbf{Extend (18)} & \textbf{Compute (25)} & \textbf{SR (83)} & \textbf{PSR (83)} \\
\midrule

\multicolumn{7}{c}{\cellcolor{gray!10}\textbf{Composite Mode}} \\ 
\midrule
\multicolumn{7}{l}{\textit{Proprietary LLMs}} \\
gpt-5                 & \heatcell{80.00}{80.00} & \heatcell{65.00}{65.00} & \heatcell{66.67}{66.67} & \heatcell{44.00}{44.00} & \heatcell{62.65}{62.65} & \heatcell{68.98}{68.98} \\
gemini-2.5-pro        & \heatcell{90.00}{90.00} & \heatcell{75.00}{\textbf{75.00}} & \heatcell{72.22}{72.22} & \heatcell{56.00}{56.00} & \heatcell{72.29}{72.29} & \heatcell{79.88}{79.88} \\
claude-sonnet-4.5     & \heatcell{90.00}{90.00} & \heatcell{75.00}{\textbf{75.00}} & \heatcell{83.33}{\textbf{83.33}} & \heatcell{68.00}{\textbf{68.00}} & \heatcell{78.31}{\textbf{78.31}} & \heatcell{82.81}{\textbf{82.81}} \\
\addlinespace 
\multicolumn{7}{l}{\textit{Open-Source LLMs}} \\
deepseek-v3.2              & \heatcell{75.00}{75.00} & \heatcell{75.00}{\textbf{75.00}} & \heatcell{66.67}{66.67} & \heatcell{36.00}{36.00} & \heatcell{61.45}{61.45} & \heatcell{68.80}{68.80} \\
Qwen3-Coder-30B-A3B-Instruct  & \heatcell{10.00}{10.00}  & \heatcell{30.00}{30.00}  & \heatcell{0.00}{0.00}  & \heatcell{0.00}{0.00}  & \heatcell{9.64}{9.64}  & \heatcell{20.54}{20.54} \\
Qwen2.5-Coder-7B-Instruct  & \heatcell{0.00}{0.00}  & \heatcell{5.00}{5.00}  & \heatcell{0.00}{0.00}  & \heatcell{0.00}{0.00}  & \heatcell{1.20}{1.20}  & \heatcell{12.09}{12.09} \\
\addlinespace
\multicolumn{7}{l}{\textit{Agentic Frameworks}} \\
AWM (gpt-5)                & \heatcell{95.00}{\textbf{95.00}} & \heatcell{65.00}{65.00} & \heatcell{72.22}{72.22} & \heatcell{36.00}{36.00} & \heatcell{65.06}{65.06} & \heatcell{71.18}{71.18} \\
AWM (gemini-2.5-pro)       & \heatcell{90.00}{90.00} & \heatcell{70.00}{70.00} & \heatcell{61.11}{61.11} & \heatcell{48.00}{48.00} & \heatcell{66.27}{66.27} & \heatcell{74.34}{74.34} \\
AWM (claude-sonnet-4.5)    & \heatcell{90.00}{90.00} & \heatcell{55.00}{55.00} & \heatcell{55.56}{55.56} & \heatcell{56.00}{56.00} & \heatcell{63.86}{63.86} & \heatcell{75.88}{75.88} \\

\midrule
\multicolumn{7}{c}{\cellcolor{gray!10}\textbf{Primitive Mode}} \\
\midrule
\multicolumn{7}{l}{\textit{Proprietary LLMs}} \\
gpt-5                 & \heatcell{15.00}{15.00} & \heatcell{40.00}{\textbf{40.00}} & \heatcell{0.00}{0.00} & \heatcell{0.00}{0.00} & \heatcell{13.25}{13.25} & \heatcell{28.27}{\textbf{28.27}} \\
gemini-2.5-pro        & \heatcell{5.00}{5.00}  & \heatcell{20.00}{20.00} & \heatcell{0.00}{0.00} & \heatcell{0.00}{0.00} & \heatcell{6.02}{6.02}  & \heatcell{19.96}{19.96} \\
claude-sonnet-4.5     & \heatcell{10.00}{10.00} & \heatcell{25.00}{25.00} & \heatcell{0.00}{0.00} & \heatcell{0.00}{0.00} & \heatcell{8.43}{8.43}  & \heatcell{21.00}{21.00} \\
\addlinespace
\multicolumn{7}{l}{\textit{Open-Source LLMs}} \\
Qwen3-VL-32B-Instruct & \heatcell{0.00}{0.00} & \heatcell{10.00}{10.00} & \heatcell{0.00}{0.00} & \heatcell{0.00}{0.00} & \heatcell{2.41}{2.41} & \heatcell{12.29}{12.29} \\
UI-TARS-1.5-7B        & \heatcell{0.00}{0.00} & \heatcell{0.00}{0.00} & \heatcell{0.00}{0.00} & \heatcell{0.00}{0.00} & \heatcell{0.00}{0.00} & \heatcell{9.88}{9.88} \\
\addlinespace
\multicolumn{7}{l}{\textit{Agentic Frameworks}} \\
AWM (gpt-5)                 & \heatcell{5.00}{5.00}  & \heatcell{40.00}{\textbf{40.00}} & \heatcell{0.00}{0.00} & \heatcell{0.00}{0.00} & \heatcell{10.84}{10.84} & \heatcell{25.82}{25.82} \\
AWM (gemini-2.5-pro)        & \heatcell{0.00}{0.00}  & \heatcell{20.00}{20.00} & \heatcell{0.00}{0.00} & \heatcell{0.00}{0.00} & \heatcell{4.82}{4.82}  & \heatcell{15.48}{15.48} \\
AWM (claude-sonnet-4.5)     & \heatcell{25.00}{\textbf{25.00}} & \heatcell{30.00}{30.00} & \heatcell{0.00}{0.00} & \heatcell{4.00}{\textbf{4.00}} & \heatcell{14.46}{\textbf{14.46}} & \heatcell{27.95}{27.95} \\
\addlinespace[2pt]
Agent S2 (gpt-5)            & \heatcell{5.00}{5.00}  & \heatcell{20.00}{20.00} & \heatcell{0.00}{0.00} & \heatcell{0.00}{0.00} & \heatcell{6.02}{6.02}  & \heatcell{18.03}{18.03} \\
Agent S2 (gemini-2.5-pro)   & \heatcell{5.00}{5.00}  & \heatcell{25.00}{25.00} & \heatcell{0.00}{0.00} & \heatcell{0.00}{0.00} & \heatcell{7.23}{7.23}  & \heatcell{18.45}{18.45} \\
Agent S2 (claude-sonnet-4.5)& \heatcell{10.00}{10.00} & \heatcell{20.00}{20.00} & \heatcell{0.00}{0.00} & \heatcell{0.00}{0.00} & \heatcell{7.23}{7.23}  & \heatcell{19.40}{19.40} \\
\bottomrule
\end{tabular}
\end{table*}

\section{Experiments}
 
We evaluate two categories of GUI agents under both modes.
For foundation models, we benchmark proprietary models (GPT-5~\citep{openai2025gpt5}, Claude-Sonnet-4.5~\citep{anthropic2025sonnet45}, Gemini-2.5-Pro~\citep{google2025gemini25pro}) as upper bounds, and open-source models including reasoning specialists (DeepSeek-V3.2~\citep{deepseekai2025deepseekv32pushingfrontieropen}, Qwen2.5-Coder-7B-Instruct~\citep{huiQwen25CoderTechnicalReport2024}) and vision specialists (UI-TARS-1.5-7B~\citep{bytedance2025uitars}, Qwen3-VL-32B-Instruct~\citep{bai2025qwen3vltechnicalreport}) as primary evaluation targets for composite and primitive modes, respectively.
For agentic frameworks, we benchmark Agent-S2~\citep{agasheAgentS2Compositional2025a}\footnote{Agent-S2 is evaluated only in primitive mode due to its dedicated design for visual grounding.}, a hierarchical agent, and AWM~\citep{wangAgentWorkflowMemory2024a}, which uses memory-augmented approaches for workflow reuse. The complete system prompts for both interaction modes are provided in Appendix~\ref{app:system_prompts}.
We also conduct an ablation on observation representations of primitive mode, reported in Appendix~\ref{app:ablation_observation}.

We host a local Scratch GUI~\citep{scratchgui} orchestrated by a unified backend server that interfaces with the browser via Playwright.
All experiments were conducted on an Ubuntu 22.04 server equipped with dual Intel Xeon Platinum 8280L CPUs, 256GB RAM, and 9 NVIDIA RTX 2080 Ti GPUs.
This server acts as a bridge, exposing a custom API to handle interactions for both modes.
In primitive mode, we default the viewport to $1280 \times 720$ and use Playwright for I/O operations. To assist visual grounding, we construct a concise list of UI elements by combining DOM elements (queried via CSS selectors) with OCR text regions from PaddleOCR~\citep{cui2025paddleocr}, and deduplicate them based on spatial overlap (see Appendix~\ref{app:element-list-fusion} for details).
In composite mode, the server withholds visual feedback and instead exposes a custom API through which both observations (pseudocode) and actions (high-level block operations) are retrieved and executed via \texttt{Scratch VM}.

We present the experimental results to answer the following research questions:
\textbf{RQ1:} How do SOTA models perform on Scratch tasks, and how does the interaction mode impact performance?
\textbf{RQ2:} Are drag-and-drop operations inherently difficult, even at the single-step level?
\textbf{RQ3:} Do perception deficits account for the drag-and-drop failures?

\subsection{RQ1: Overall Performance}

Table~\ref{tab:results} summarizes the performance of selected baselines across four task categories and two interaction modes. The results highlight critical insights regarding the capabilities and limitations of current base models and agents within the Scratch environment.

\noindent\textbf{Key Finding 1 (Reasoning--Acting Gap).}
In composite mode, Claude-Sonnet-4.5 achieves 78.31\% SR, while the open-source DeepSeek-v3.2 attains 61.45\%, demonstrating that contemporary LLMs possess strong logical reasoning capabilities. In stark contrast, performance drops to 14.46\% in primitive mode (AWM with Claude-Sonnet-4.5), revealing severe bottlenecks in executing plans through fine-grained drag-and-drop operations despite competent high-level planning.

\noindent\textbf{Key Finding 2 (Category Sensitivity).}
The magnitude of performance degradation varies significantly depending on the task nature. Create tasks suffer the most precipitous decline; while their high success rates in composite mode substantiate the sufficiency of the models' logical capabilities, the necessity for extensive drag-and-drop actions in primitive mode introduces a severe visuomotor bottleneck. Debug tasks are more resilient because they typically involve localized block adjustments and parameter updates, with fewer high-precision dragging operations than Create tasks. Finally, Extend and Compute tasks exhibit consistently lower success rates across both modes due to the combined requirement of strong logical reasoning and extensive drag-and-drop interactions.

\noindent\textbf{Key Finding 3 (Framework Limitations).}
We further examine why two GUI agentic frameworks that performed well in prior benchmarks did not yield performance gains in this study, and instead underperformed relative to naive agents. For the \textbf{AWM} framework, effectiveness depends on inducing workflows from successful task executions to improve the success rate of subsequent tasks. However, in primitive mode, the high failure rate of drag-and-drop operations prevented AWM from accumulating a sufficient number of successful trajectories. This cold start limitation not only hindered the workflow summarization mechanism from functioning as intended, but also could have introduced noise that interfered with the decision-making process, thereby degrading performance. With respect to \textbf{Agent-S2}, it adopts a collaborative architecture in which a strong LLM performs reasoning and planning while a smaller model performs visual localization. As the capabilities of the base model increase, the specialized smaller model no longer provides a clear advantage for visual localization. The experimental results indicate that, after applying this framework, the success rates of GPT-5 and Claude-Sonnet-4.5, which exhibit stronger visual localization capability, decreased, whereas Gemini-2.5-Pro, which exhibits comparatively weaker visual localization capability, achieved improved performance under this architecture.

\subsection{RQ2: Drag-and-Drop Precision Analysis}

\def\rqTwoHeatMaxMix{55} 
\def\rqTwoHeatBoxW{9.0mm}

\newcommand{\rqTwoHeatHiRange}[4]{%
  \begingroup
  \pgfmathsetmacro{\p}{max(0,min(1,(#1-#2)/(#3-#2)))}
  \pgfmathsetmacro{\mix}{abs(\p-0.5)/0.5*\rqTwoHeatMaxMix}
  \ifdim \p pt<0.5pt
    \def\rqTwoHeatCol{red!\mix!white}%
  \else
    \def\rqTwoHeatCol{green!\mix!white}%
  \fi
  \tikz[baseline=(X.base)]\node[
    fill=\rqTwoHeatCol,
    rounded corners=1.2pt,
    inner xsep=0pt,
    inner ysep=0.8pt,
    minimum width=\rqTwoHeatBoxW,
    text width=\rqTwoHeatBoxW,
    align=center
  ](X){#4};%
  \endgroup
}
\newcommand{\rqTwoHeatLoRange}[4]{%
  \begingroup
  \pgfmathsetmacro{\p}{max(0,min(1,(#1-#2)/(#3-#2)))}
  \pgfmathsetmacro{\pinv}{1-\p}
  \pgfmathsetmacro{\mix}{abs(\pinv-0.5)/0.5*\rqTwoHeatMaxMix}
  \ifdim \pinv pt<0.5pt
    \def\rqTwoHeatCol{red!\mix!white}%
  \else
    \def\rqTwoHeatCol{green!\mix!white}%
  \fi
  \tikz[baseline=(X.base)]\node[
    fill=\rqTwoHeatCol,
    rounded corners=1.2pt,
    inner xsep=0pt,
    inner ysep=0.8pt,
    minimum width=\rqTwoHeatBoxW,
    text width=\rqTwoHeatBoxW,
    align=center
  ](X){#4};%
  \endgroup
}

\def\rqTwoSRMin{0}
\def\rqTwoSRMax{55}
\def\rqTwoAccMin{0}
\def\rqTwoAccMax{100}
\def\rqTwoErrMin{0}
\def\rqTwoErrMax{330}

\newcommand{\rqTwoSR}[2]{\rqTwoHeatHiRange{#1}{\rqTwoSRMin}{\rqTwoSRMax}{#2}}
\newcommand{\rqTwoAcc}[2]{\rqTwoHeatHiRange{#1}{\rqTwoAccMin}{\rqTwoAccMax}{#2}}
\newcommand{\rqTwoErr}[2]{\rqTwoHeatLoRange{#1}{\rqTwoErrMin}{\rqTwoErrMax}{#2}}

\begin{table*}[t]
    \centering
      \caption{Single-Step Drag Benchmark results on 60 atomic drag-and-drop tasks. The table summarizes success rates (Pass@k), component-wise pass rates, and spatial coordinate errors. See Appendix~\ref{app:drag_category_results} for results by interaction type.}
    \label{tab:rq2_results_all}
    \begin{tabular}{llccccccc}
        \toprule
        & & \multicolumn{3}{c}{Success Rate (\%)} & \multicolumn{2}{c}{Component Acc. (\%)} & \multicolumn{2}{c}{Spatial Error (px)} \\
        \cmidrule(lr){3-5} \cmidrule(lr){6-7} \cmidrule(lr){8-9}
        Model & Setting & @1 & @2 & @3 & Start & End & Start & End \\
        \midrule
        \multirow{3}{*}{GPT-5} 
          & Baseline  & \rqTwoSR{23.33}{23.33} & \rqTwoSR{28.33}{28.33} & \rqTwoSR{35.00}{35.00} & \rqTwoAcc{66.67}{66.67} & \rqTwoAcc{32.50}{32.50} & \rqTwoErr{28.32}{28.32} & \rqTwoErr{36.05}{36.05} \\
          & GT Start  & \rqTwoSR{26.67}{26.67} & \rqTwoSR{36.67}{36.67} & \rqTwoSR{46.67}{46.67} & \rqTwoAcc{99.44}{99.44} & \rqTwoAcc{30.17}{30.17} & \rqTwoErr{60.00}{60.00} & \rqTwoErr{39.06}{39.06} \\
          & Knowledge & \rqTwoSR{31.67}{31.67} & \rqTwoSR{41.67}{\textbf{41.67}} & \rqTwoSR{51.67}{\textbf{51.67}} & \rqTwoAcc{63.33}{63.33} & \rqTwoAcc{43.86}{\textbf{43.86}} & \rqTwoErr{13.01}{\textbf{13.01}} & \rqTwoErr{26.56}{\textbf{26.56}} \\
        \midrule
        \multirow{3}{*}{Qwen3-VL-32B-Instruct} 
          & Baseline  & \rqTwoSR{26.67}{26.67} & \rqTwoSR{31.67}{31.67} & \rqTwoSR{33.33}{33.33} & \rqTwoAcc{67.78}{67.78} & \rqTwoAcc{36.89}{36.89} & \rqTwoErr{142.79}{142.79} & \rqTwoErr{45.68}{45.68} \\
          & GT Start  & \rqTwoSR{35.00}{\textbf{35.00}} & \rqTwoSR{38.33}{38.33} & \rqTwoSR{40.00}{40.00} & \rqTwoAcc{100.00}{\textbf{100.00}} & \rqTwoAcc{32.22}{32.22} & \textemdash & \rqTwoErr{43.82}{43.82} \\
          & Knowledge & \rqTwoSR{28.33}{28.33} & \rqTwoSR{40.00}{40.00} & \rqTwoSR{51.67}{\textbf{51.67}} & \rqTwoAcc{67.78}{67.78} & \rqTwoAcc{41.80}{41.80} & \rqTwoErr{75.75}{75.75} & \rqTwoErr{34.47}{34.47} \\
        \midrule
        \multirow{3}{*}{UI-TARS-1.5-7B} 
          & Baseline  & \rqTwoSR{1.67}{1.67} & \rqTwoSR{6.67}{6.67} & \rqTwoSR{6.67}{6.67} & \rqTwoAcc{49.44}{49.44} & \rqTwoAcc{5.62}{5.62} & \rqTwoErr{238.50}{238.50} & \rqTwoErr{84.58}{84.58} \\
          & GT Start  & \rqTwoSR{13.33}{13.33} & \rqTwoSR{13.33}{13.33} & \rqTwoSR{13.33}{13.33} & \rqTwoAcc{56.67}{56.67} & \rqTwoAcc{23.53}{23.53} & \rqTwoErr{247.46}{247.46} & \rqTwoErr{71.89}{71.89} \\
          & Knowledge & \rqTwoSR{16.67}{16.67} & \rqTwoSR{16.67}{16.67} & \rqTwoSR{18.33}{18.33} & \rqTwoAcc{45.00}{45.00} & \rqTwoAcc{37.04}{37.04} & \rqTwoErr{250.66}{250.66} & \rqTwoErr{90.34}{90.34} \\
        \bottomrule
    \end{tabular}%
\end{table*}

In this subsection, we explain the sharp performance drop observed in primitive mode by isolating models' fine-grained drag-and-drop capability. Building on the 83 tasks in \bench, we construct a \textbf{Single-Step Drag Benchmark} with 60 atomic tasks that each can be completed with a single drag action, covering two interaction scenarios in Scratch: \textit{Direct Connection} and \textit{Slot Insertion} (see examples in Appendix \ref{app:drag_examples}). We evaluate GPT-5, Qwen3-VL-32B-Instruct, and UI-TARS-1.5-7B under three settings with increasing task-side assistance: \textit{Baseline} (raw screenshot), \textit{GT Start} (ground-truth start position), and \textit{Knowledge} (heuristic rules; Appendix \ref{app:heuristics-for-drag-and-drop}). To quantify operation errors, we compute ground-truth feasible regions for valid start/end positions via a BFS-based algorithm (Appendix \ref{app:bfs-feasible-regions}) and define pixel-level \textit{Spatial Error} as the Euclidean distance from predicted coordinates to the nearest feasible point when predictions fall outside the feasible region.

\noindent\textbf{Key Finding 4 (Single-Step Success is Still Low).}
Table \ref{tab:rq2_results_all} shows that even with task complexity reduced to a single drag-and-drop operation, success rates remain low across models, with the best performance (GPT-5) achieving only 23.33\% Pass@1. This directly answers a key question raised by RQ1: \textit{does the reasoning--acting gap stem from long-horizon task complexity, or from fundamental visuomotor control deficits?} The single-step benchmark isolates the latter, indicating that per-step execution precision, rather than multi-step planning, is the primary bottleneck. Consequently, in long-horizon tasks (9.14 block modifications on average), even moderate per-step failure rates compound into very low end-to-end success in primitive mode.

\noindent\textbf{Key Finding 5 (Endpoint Localization is the Dominant Failure Mode).}
Providing ground-truth start positions improves Pass@1 for all three models, but endpoint localization remains poor. Under \textit{GT Start}, Start Component Acc. becomes near-perfect for GPT-5 (99.44\%) and Qwen3-VL (100.00\%), yet End Component Acc. stays low for all three models (GPT-5: 30.17\%, Qwen3-VL: 32.22\%, UI-TARS: 23.53\%). This suggests that predicting the correct connection target (``where to drop'') is substantially harder than identifying a valid grasp point (``where to grab''), and thus increasingly constrains performance as the interaction horizon grows.

\noindent\textbf{Key Finding 6 (Heuristic Priors Help, but Do Not Fully Solve the Problem).}
Across all three models, the \textit{Knowledge} setting yields the best overall results. With heuristic grounding, Qwen3-VL-32B-Instruct matches GPT-5 on this benchmark (Pass@3: 51.67\%), suggesting that explicit grounding rules can narrow the gap between proprietary and open-source models on specific low-level operations. In addition to improving Pass@k, heuristics reduce localization errors for GPT-5 and Qwen3-VL, whereas UI-TARS shows slightly higher spatial error, indicating that smaller models do not reliably translate textual heuristics into pixel-level action refinement.

\subsection{RQ3: Visual Perception Analysis}

Having established that drag-and-drop operations frequently fail even at the single-step level (RQ2), we now investigate whether perception deficits account for these failures. If models cannot accurately perceive block structures and spatial relationships, execution failures would be expected. We construct a \textbf{Visual Perception Benchmark} that decouples perception from action by posing static visual QA on Scratch screenshots. Based on the visual demands of executing Scratch tasks, the benchmark targets three critical perception scenarios with 200 manually curated samples: detecting \textbf{block connections}, identifying \textbf{block existence under occlusion}, and reading \textbf{field values} (see Appendix~\ref{app:perception_examples}).

\noindent\textbf{Key Finding 7 (Strong Static Perception Across SOTA Models).}
Table~\ref{tab:rq3_results} shows that GPT-5 achieves 90.5\% overall accuracy, with near-ceiling performance on field reading (98.4\%). Among open models, Qwen3-VL-32B-Instruct reaches 77.5\%, outperforming UI-TARS-1.5-7B (67.5\%), especially on connection and existence checks.

\noindent\textbf{Key Finding 8 (Perception is Not the Bottleneck in Primitive Mode).}
Despite high perception accuracy, drag-and-drop success remains much lower (e.g., GPT-5: 90.5\% perception vs. 23.33\% single-step drag). This gap indicates that, for current agents, the dominant limitation lies in translating correctly perceived GUI state into precise execution.

\noindent\textbf{Key Finding 9 (Execution Errors Dominate Even Under Near-Correct Understanding).}
Taken together with RQ2, these results establish a clear diagnostic: failures in primitive mode primarily arise from control execution, particularly the fine-grained coordinate prediction required for reliable snapping and block manipulation, rather than from misreading blocks or their relationships.

\def\rqThreeHeatMaxMix{65}
\def\rqThreeHeatBoxW{7.8mm}
\def\rqThreeMin{60}
\def\rqThreeMid{80}
\def\rqThreeMax{100}
\newcommand{\rqThreeHeat}[2]{%
  \begingroup
  \pgfmathsetmacro{\praw}{#1}%
  \pgfmathsetmacro{\p}{max(\rqThreeMin,min(\rqThreeMax,\praw))}
  \ifdim \p pt<\rqThreeMid pt
    \pgfmathsetmacro{\t}{(\rqThreeMid-\p)/(\rqThreeMid-\rqThreeMin)}
    \pgfmathsetmacro{\mix}{\t*\rqThreeHeatMaxMix}%
    \def\rqThreeHeatCol{red!\mix!white}%
  \else
    \pgfmathsetmacro{\t}{(\p-\rqThreeMid)/(\rqThreeMax-\rqThreeMid)}
    \pgfmathsetmacro{\mix}{\t*\rqThreeHeatMaxMix}%
    \def\rqThreeHeatCol{green!\mix!white}%
  \fi
  \tikz[baseline=(X.base)]\node[
    fill=\rqThreeHeatCol,
    rounded corners=1.2pt,
    inner xsep=0pt,
    inner ysep=0.8pt,
    minimum width=\rqThreeHeatBoxW,
    text width=\rqThreeHeatBoxW,
    align=center
  ](X){#2};%
  \endgroup
}

\begin{table}[htb!]
  \centering
  \caption{Performance on Visual Perception Benchmark. We evaluate the following three dimensions: \textbf{Connection} (detecting whether blocks are physically connected), \textbf{Existence} (identifying blocks under occlusion or visual clutter), and \textbf{Field} (reading parameter values with equivalence checking or exact string matching). Accuracy is reported in percentages.}
  \label{tab:rq3_results}
  \resizebox{\columnwidth}{!}{
  \begin{tabular}{lcccc}
  \toprule
  Model & Connection & Existence & Field & Overall \\
  \midrule
  UI-TARS-1.5-7B & \rqThreeHeat{61.7}{61.7} & \rqThreeHeat{63.7}{63.7} & \rqThreeHeat{78.4}{78.4} & \rqThreeHeat{67.5}{67.5} \\
  Qwen3-VL-32B-Instruct & \rqThreeHeat{65.0}{65.0} & \rqThreeHeat{78.8}{78.8} & \rqThreeHeat{88.4}{88.4} & \rqThreeHeat{77.5}{77.5} \\
  GPT-5 & \rqThreeHeat{78.3}{\textbf{78.3}} & \rqThreeHeat{93.8}{\textbf{93.8}} & \rqThreeHeat{98.4}{\textbf{98.4}} & \rqThreeHeat{90.5}{\textbf{90.5}} \\
  \bottomrule
  \end{tabular}
  }
\end{table}

\section{Conclusion}

In this paper, we introduced \bench, the first benchmark for evaluating multimodal GUI agents on block-based programming. \bench comprises 83 tasks across four categories, utilizing a dual-mode interaction protocol to disentangle reasoning from execution. Our experiments reveal a significant reasoning--acting gap: while current models excel at high-level planning, they fail at precise drag-and-drop manipulations. Targeted diagnostics isolate endpoint localization as the dominant bottleneck, while ruling out perception deficits.

\clearpage
\newpage

\section*{Impact Statement}

\bench advances machine learning for computer science education by enabling AI agents to serve as interactive programming tutors. While this technology promises to democratize access to high-quality instruction, deploying it requires careful consideration. Developers must guard against student over-reliance, which could diminish independent problem-solving skills. Additionally, given the user demographic of Scratch, strict adherence to data privacy and content safety for minors is paramount. We emphasize that AI agents should scaffold the learning process rather than automate creativity.

\bibliography{ref}
\bibliographystyle{icml2026}

\newpage
\appendix
\onecolumn
\phantomsection
\section*{Appendix Contents}
\begin{tcolorbox}[
  colback=gray!2,
  colframe=gray!40,
  boxrule=0.6pt,
  arc=2pt,
  left=6pt,
  right=6pt,
  top=6pt,
  bottom=6pt,
]
\begin{itemize}[leftmargin=*, itemsep=2pt, topsep=2pt]
  \item \hyperref[app:ablation_observation]{\textbf{A.} Ablation Study on Observation Representations}
  \item \hyperref[sec:source_projects]{\textbf{B.} Details of Selected Source Projects}
  \item \hyperref[sec:task_examples]{\textbf{C.} Representative Task Examples}
  \item \hyperref[sec:primitive_action_space]{\textbf{D.} Detailed Action Space of Primitive Mode}
  \item \hyperref[sec:composite_action_space]{\textbf{E.} Detailed Action Space of Composite Mode}
  \item \hyperref[sec:comp_observations]{\textbf{F.} Example of Composite Mode Observation}
  \item \hyperref[app:drag_examples]{\textbf{G.} Single-Step Drag-and-Drop Task Examples}
  \item \hyperref[app:perception_examples]{\textbf{H.} Visual Perception Task Examples}
  \item \hyperref[app:drag_category_results]{\textbf{I.} Results for Single-Step Drag Benchmark in Task Categories}
  \item \hyperref[app:bfs-feasible-regions]{\textbf{J.} BFS-Based Feasible Region Computation}
  \item \hyperref[app:heuristics-for-drag-and-drop]{\textbf{K.} Heuristics Hints for Drag-and-Drop Tasks}
  \item \hyperref[sec:serialization_example]{\textbf{L.} Project Serialization Example}
  \item \hyperref[sec:element_list_example]{\textbf{M.} Example of Element List Observation}
  \item \hyperref[sec:eval_script_example]{\textbf{N.} Example of Evaluation Script}
  \item \hyperref[sec:appendix_prompts]{\textbf{O.} Prompts for Semi-Automated Generation}
  \begin{itemize}[leftmargin=1.6em, itemsep=1pt, topsep=1pt]
    \item \hyperref[app:prompt_create]{O.1 Task Idea Generation: Create}
    \item \hyperref[app:prompt_debug_extend]{O.2 Task Idea Generation: Debug and Extend}
    \item \hyperref[app:prompt_eval]{O.3 Evaluation Script Generation}
  \end{itemize}
  \item \hyperref[app:system_prompts]{\textbf{P.} System Prompts for Two Interaction Modes}
  \begin{itemize}[leftmargin=1.6em, itemsep=1pt, topsep=1pt]
    \item \hyperref[app:system_prompt_primitive]{P.1 Primitive Mode System Prompt}
    \item \hyperref[app:system_prompt_composite]{P.2 Composite Mode System Prompt}
  \end{itemize}
  \item \hyperref[app:element-list-fusion]{\textbf{Q.} Element List Construction and OCR--DOM Fusion}
  \item \hyperref[sec:llm]{\textbf{R.} Use of LLMs}
\end{itemize}
\end{tcolorbox}

\clearpage


\section{Ablation Study on Observation Representations}
\label{app:ablation_observation}
In the primary experimental design of this study, the primitive mode provided agents with screenshots and a structured element list as the observation space.  An analysis of failure trajectories identified a systematic interaction defect in which agents frequently relied on indices from the element list to execute drag-and-drop operations. This behavior led to imprecise positioning because index-based placement consistently targeted the geometric center of the destination block. Although the Scratch environment supports automatic snapping, placement at the geometric center often fails to coincide with the specific connection regions required to activate this mechanism. Aiming at preserve a clear distinction from the composite mode, which implicitly incorporates such low-level optimizations, we didn't introduce heuristic offsets to facilitate snapping.

To further examine the trade-off between the semantic indexing support provided by the element list and its propensity to induce imprecise interaction behaviors, an ablation study was conducted. In this setting, the element list was removed, and models were required to complete tasks based solely on screenshots. Experiments were conducted on 24 representative samples spanning four task categories. As reported in Table \ref{tab:rq_4_results}, the removal of textual modality support resulted in a reduction in success rates across both models, with GPT-5 decreasing from 16.67 percent to 8.33 percent. These results indicate that, despite challenges related to execution-level alignment, structured textual observations remain essential auxiliary information for enabling models to interpret complex graphical user interface environments, as current multimodal models continue to exhibit limitations in accurate program construction based solely on visual input.

\begin{table}[htb!]
    \centering
    \caption{Ablation study on observation representations in primitive mode. We compare agent performance when provided with \textbf{Screenshots only} versus \textbf{Element List + Screenshot} (visual input augmented with textual element metadata). Results are reported on 24 representative samples across four task categories. Success rates are in percentages (\%). Despite challenges with index-based positioning, structured textual observations remain essential for GUI understanding.}
    \label{tab:rq_4_results}
    \begin{tabular}{llccccc}
        \toprule
        Model & Context Mode & Overall & Create & Debug & Extend & Compute \\
        \midrule
        \multirow{2}{*}{GPT-5} 
          & Screenshots & 8.33 & 0.00 & 33.33 & 0.00 & 0.00 \\
          & Element List + Screenshot & 16.67 & 33.33 & 33.33 & 0.00 & 0.00 \\
        \midrule
        \multirow{2}{*}{Gemini-2.5-Pro} 
          & Screenshots & 0.00 & 0.00 & 0.00 & 0.00 & 0.00 \\
          & Element List + Screenshot & 8.33 & 0.00 & 33.33 & 0.00 & 0.00 \\
        \bottomrule
    \end{tabular}%
\end{table}

\section{Details of Selected Source Projects}
\label{sec:source_projects}

In this section, we list the 8 official Scratch starter projects that served as the foundation for the Debug and Extend tasks in \bench. These projects were selected to ensure coverage of diverse programming concepts (e.g., broadcasting, cloning, physics simulation) and project categories (Games, Animations, etc.). See Table~\ref{tab:starter_projects} for the complete list and key concepts.

\begin{table}[htb!]
    \centering
    \caption{Eight official Scratch starter projects used to seed Debug and Extend tasks in \bench, with derived task counts and key concepts.}
    \label{tab:starter_projects}
    \begin{tabular}{l c p{6.5cm}}
    \toprule
    \textbf{Project Name} & \textbf{Tasks Derived} & \textbf{Key Concepts} \\ \midrule
    \href{https://scratch.mit.edu/projects/1106739913/}{X and Y Coordinates} & 2 & Coordinates, Input, Debugging \\
    \href{https://scratch.mit.edu/projects/1110545496/}{Make It Fly} & 5 & Gravity, Scrolling, Events \\
    \href{https://scratch.mit.edu/projects/1106220358/}{Math Game} & 4 & Arithmetic, Broadcasting, Variables \\
    \href{https://scratch.mit.edu/projects/10128431/}{Maze Starter} & 10 & Collision detection, Motion, Control \\
    \href{https://scratch.mit.edu/projects/1105118803/}{Make a Mouse Trail} & 3 & Cloning, Mouse Sensing, Effects \\
    \href{https://scratch.mit.edu/projects/10128067/}{Dance Party} & 1 & Costumes, Sound, Synchronization \\
    \href{https://scratch.mit.edu/projects/10128515/}{Pong Starter} & 5 & Physics loop, Collision detection \\
    \href{https://scratch.mit.edu/projects/1106279050/}{Simple Circuit Simulation} & 8 & Logic, States, Interactive UI \\
    \bottomrule
    \end{tabular}
    
    {\small \textit{Note: Click on the project names to visit the official Scratch pages.}}
\end{table}

\clearpage

\section{Representative Task Examples}
\label{sec:task_examples}

In this section, we provide detailed examples for each of the four task categories in \bench: Create, Debug, Extend, and Compute. For each category, we showcase two representative tasks, including their instructions, initial states, golden solutions, and evaluation test cases.

\begin{longtable}{|p{0.95\linewidth}|}
\hline
\multicolumn{1}{|c|}{\textbf{Category 1: Create Tasks}} \\
\hline
\textbf{Type:} Create \\
\textbf{Instruction:} Starting from an empty Scratch project with a single default sprite named 'balloon', you should complete the following: 1) Create a variable named 'score' and initialize it to 0 when the green flag is clicked. 2) Each time the balloon is clicked, increase 'score' by 1 and immediately move the balloon to a random position on the stage. \\
\hline
\vspace{2pt}
\begin{minipage}{0.48\linewidth}
\centering
\includegraphics[width=\linewidth]{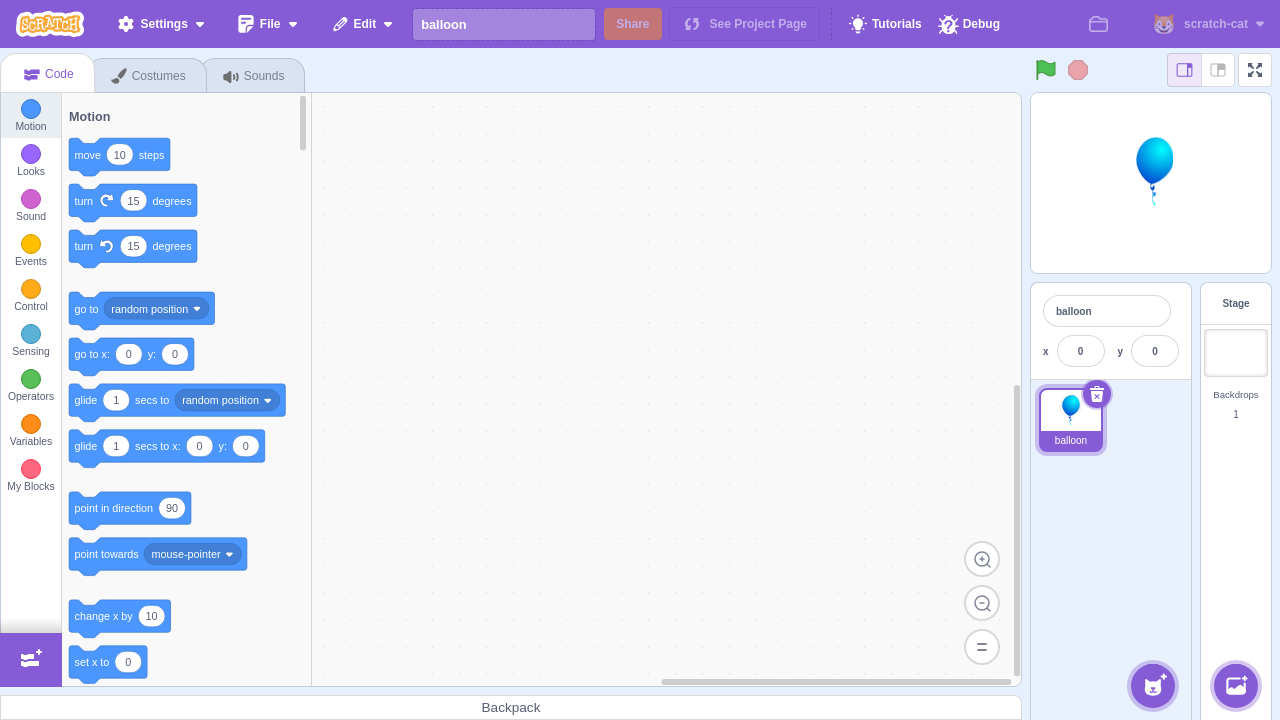} 
\centerline{\small (a) Initial Project State}
\end{minipage}
\hfill
\begin{minipage}{0.48\linewidth}
\centering
\includegraphics[width=\linewidth]{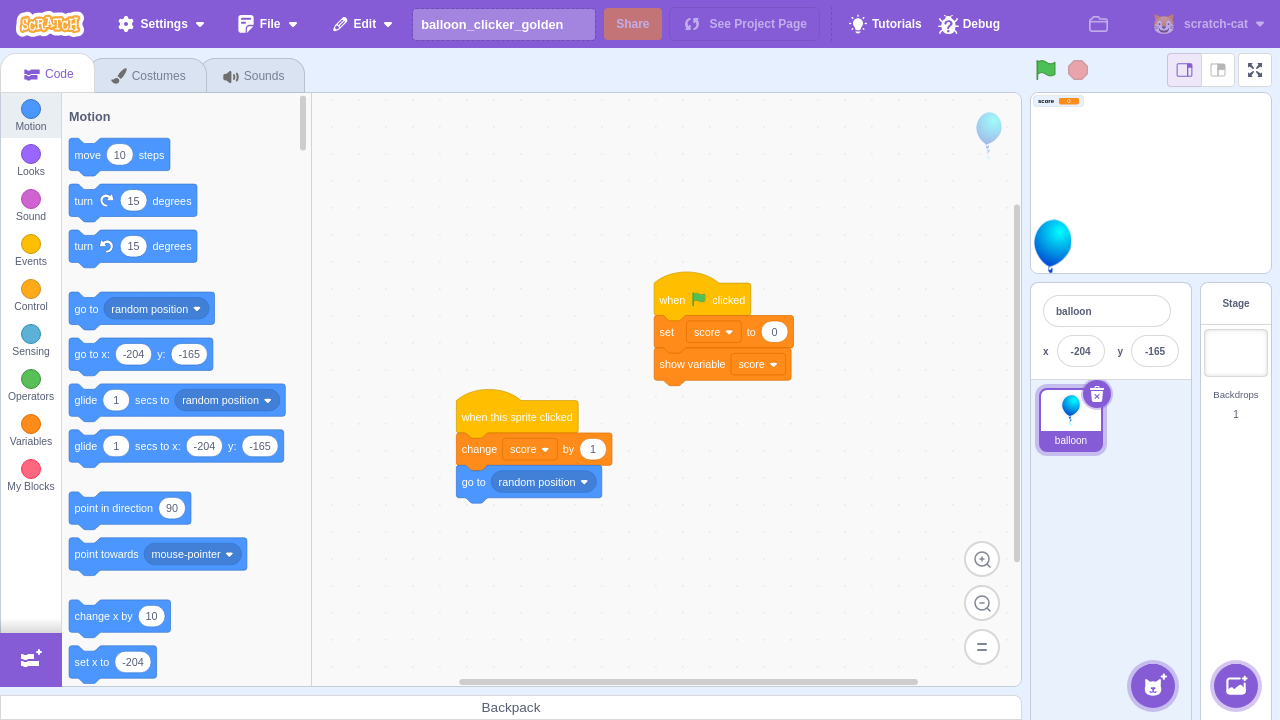}
\centerline{\small (b) Golden Project State}
\end{minipage} 
\vspace{2pt} \\
\hline
\textbf{Test Cases:} 1) Verify 'balloon' sprite and 'score' variable exist. 2) Click 'balloon' 3 times: verify 'score' increments each time. 3) Verify 'balloon' position changes (random jump) after each click. \\
\hline
\hline
\textbf{Type:} Create \\
\textbf{Instruction:} Starting from an empty Scratch project with a single default sprite named 'jellyfish', you should complete the following: 1) When the green flag is clicked, hide the jellyfish sprite. 2) Continuously create clones of the jellyfish at random X positions along the bottom. 3) For each clone, make it rise upward at a constant speed until it reaches the top edge, then delete the clone. \\
\hline
\vspace{2pt}
\begin{minipage}{0.48\linewidth}
\centering
\includegraphics[width=\linewidth]{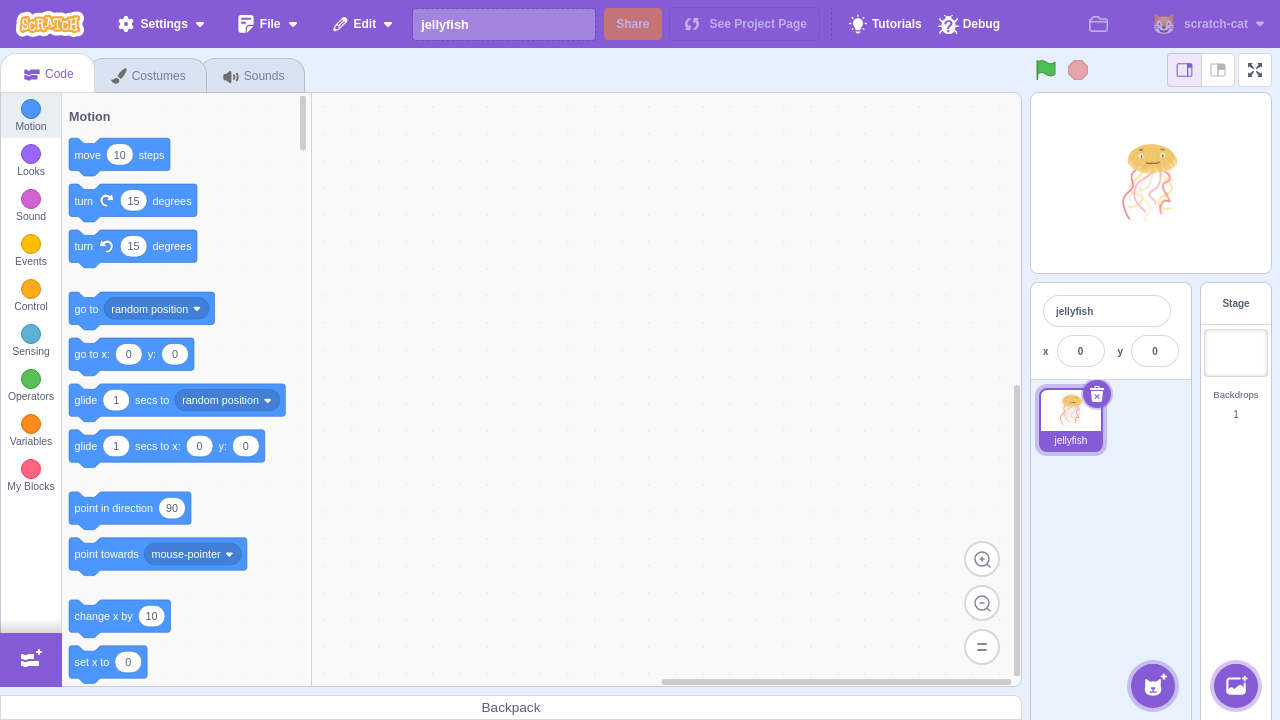} 
\centerline{\small (a) Initial Project State}
\end{minipage}
\hfill
\begin{minipage}{0.48\linewidth}
\centering
\includegraphics[width=\linewidth]{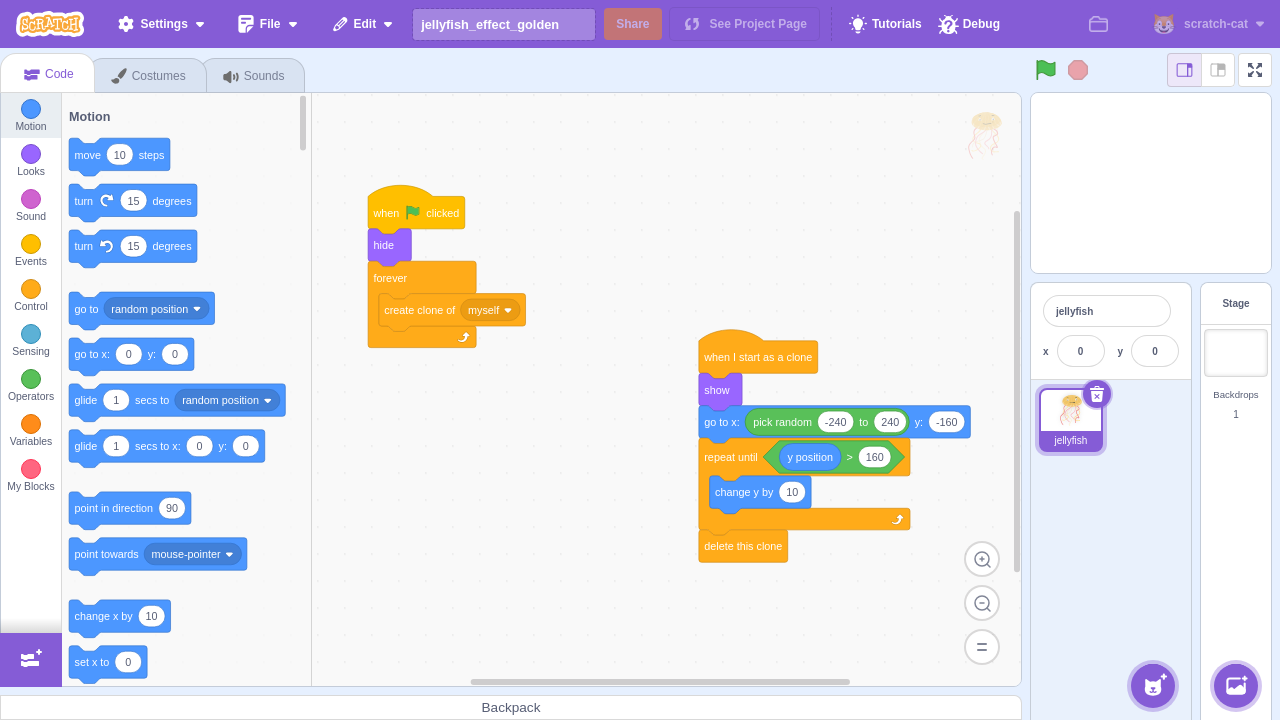}
\centerline{\small (b) Golden Project State}
\end{minipage} 
\vspace{2pt} \\
\hline
\textbf{Test Cases:} 1) Verify 'jellyfish' sprite exists. 2) Creation: At least 3 clones created periodically. 3) Motion: At least 2 clones spawn near bottom and rise $\ge$ 50px. 4) Cleanup: At least 1 clone deleted near top edge. 5) Randomness: Clones spawn at different X coordinates. \\
\hline

\hline
\multicolumn{1}{|c|}{\textbf{Category 2: Debug Tasks}} \\
\hline
\textbf{Type:} Debug \\
\textbf{Instruction:} This is a basic Pong-style game where a ball bounces around the screen and players control a paddle that follows the mouse. The ball changes direction when hitting the paddle and the game ends when the ball touches red areas. But now in this Pong Game, the Paddle's movement is inconsistent with the mouse movement. Please help me fix this issue. \\
\hline
\vspace{2pt}
\begin{minipage}{0.48\linewidth}
\centering
\includegraphics[width=\linewidth]{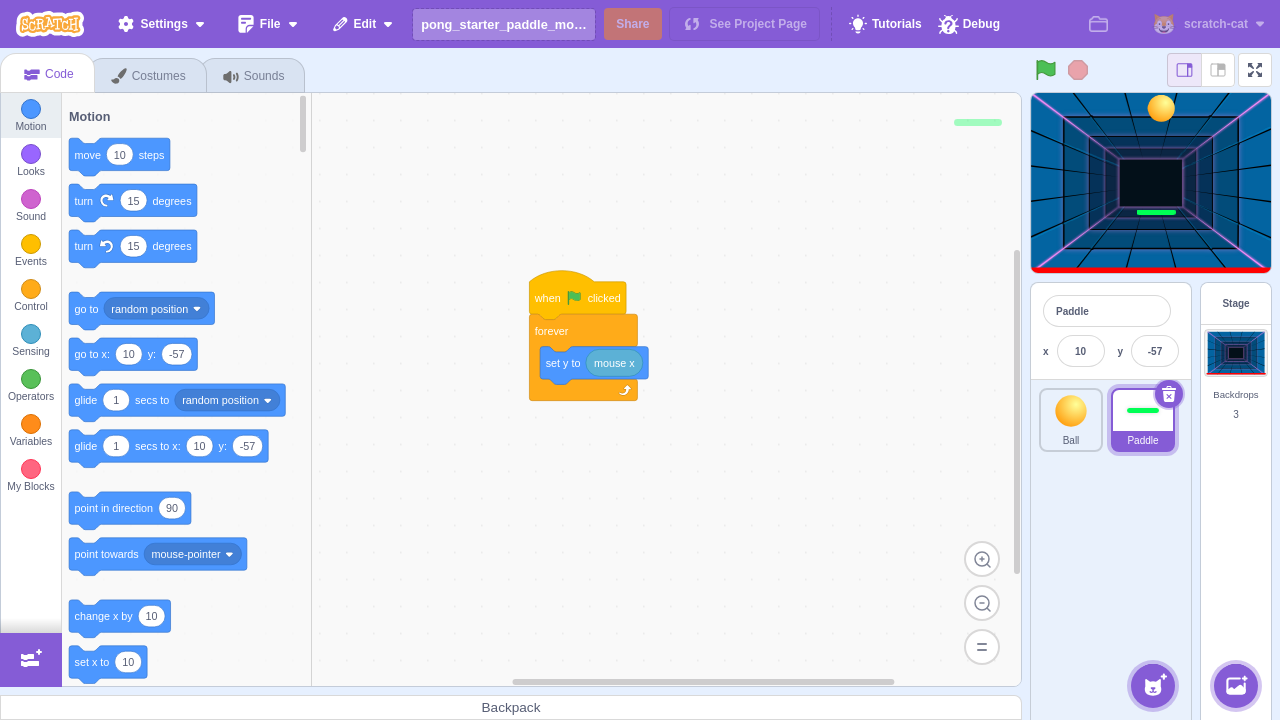} 
\centerline{\small (a) Initial Project State}
\end{minipage}
\hfill
\begin{minipage}{0.48\linewidth}
\centering
\includegraphics[width=\linewidth]{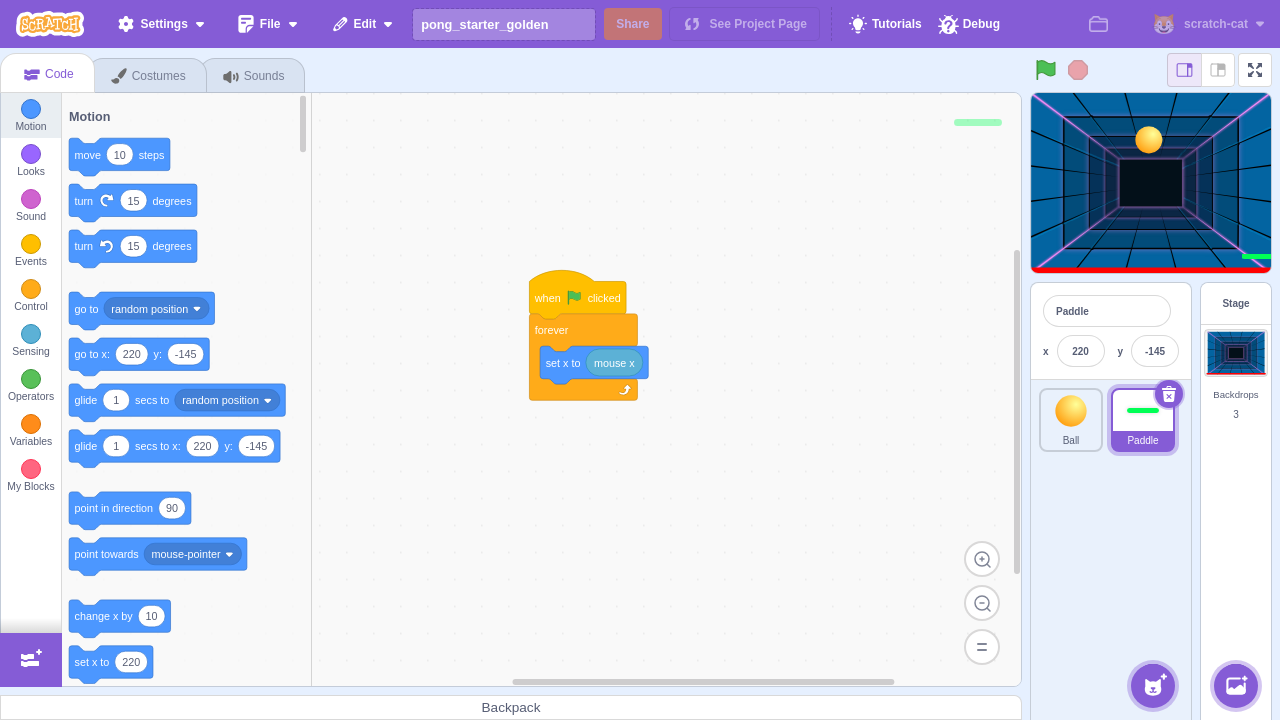}
\centerline{\small (b) Golden Project State}
\end{minipage} 
\vspace{2pt} \\
\hline
\textbf{Test Cases:} 1) Verify 'Paddle' sprite exists. 2) Move mouse to x = -200, -100, 0, 100, 200: Verify Paddle's x-position follows within 10px tolerance. \\
\hline
\hline
\textbf{Type:} Debug \\
\textbf{Instruction:} This is an interactive circuit simulation where users can drag and connect a battery, switch, and lightbulb to complete an electrical circuit. When all components are properly connected and the switch is turned on, the lightbulb illuminates with sound effects. But now in this Simple Circuit project, even when 'ON or OFF' is ON and 'connections complete' equals 3, the Lightbulb does not turn on and remains in its OFF visual state. Please help me fix this issue so that the lightbulb turns on when the circuit is complete and the switch is ON. \\
\hline
\vspace{2pt}
\begin{minipage}{0.48\linewidth}
\centering
\includegraphics[width=\linewidth]{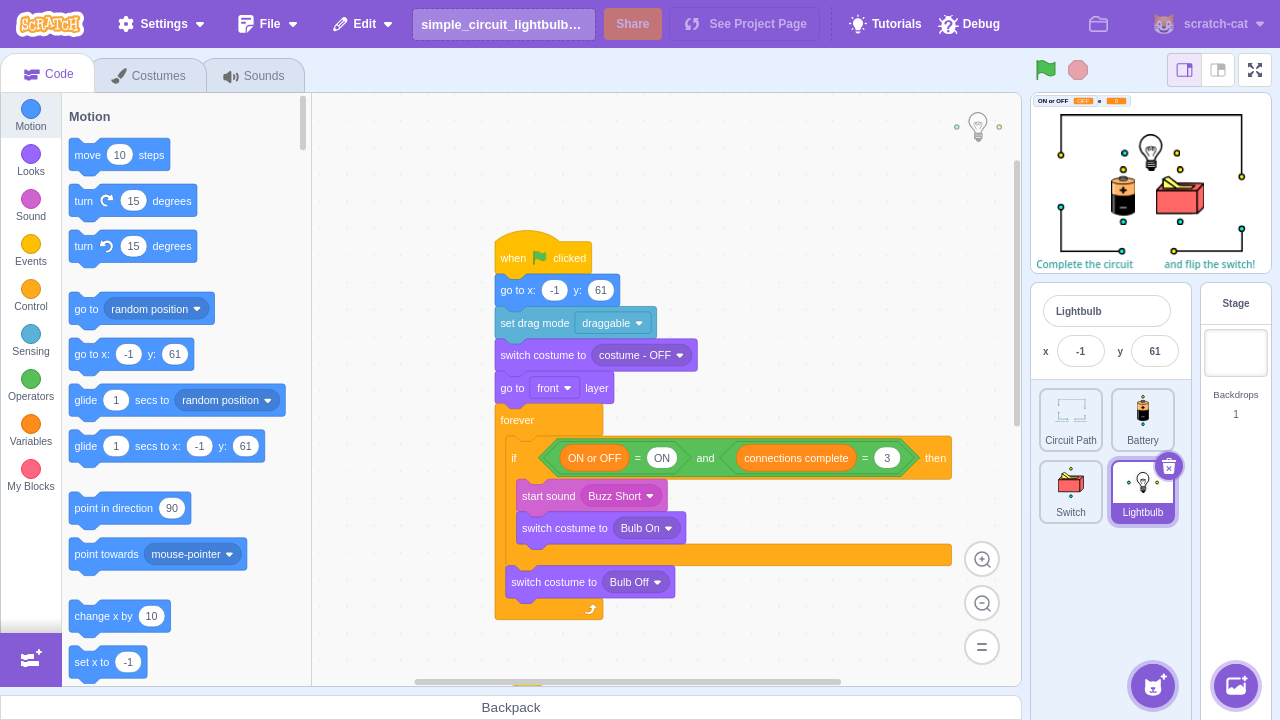} 
\centerline{\small (a) Initial Project State}
\end{minipage}
\hfill
\begin{minipage}{0.48\linewidth}
\centering
\includegraphics[width=\linewidth]{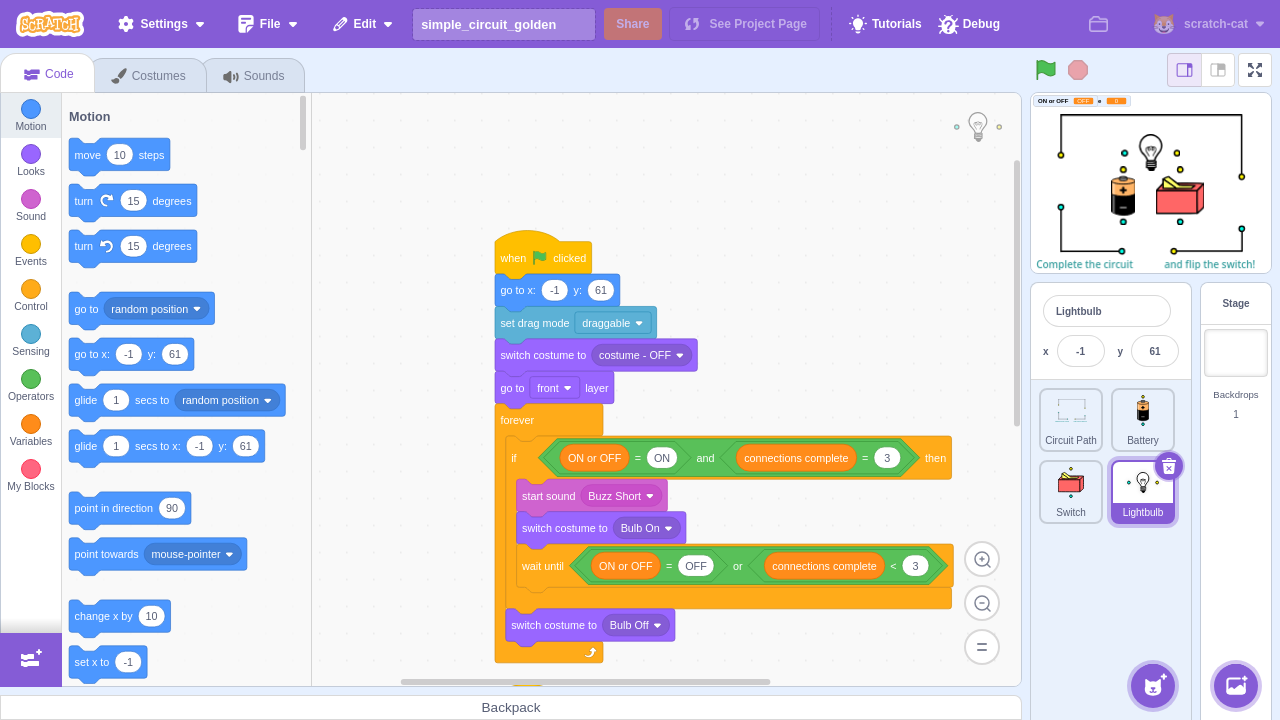}
\centerline{\small (b) Golden Project State}
\end{minipage} 
\vspace{2pt} \\
\hline
\textbf{Test Cases:} 1) Set 'ON or OFF'=ON, 'connections'=2 $\rightarrow$ Check Lightbulb stays OFF. 2) Set 'ON or OFF'=ON, 'connections'=3 $\rightarrow$ Check Lightbulb turns ON. 3) Set 'ON or OFF'=OFF, 'connections'=3 $\rightarrow$ Check Lightbulb stays OFF. \\
\hline

\hline
\multicolumn{1}{|c|}{\textbf{Category 3: Extend Tasks}} \\
\hline
\textbf{Type:} Extend \\
\textbf{Instruction:} This is a simple flying game where a cat sprite can be controlled with arrow keys to fly up and down while buildings scroll continuously across the screen from right to left, creating a side-scrolling flight experience. We want to add a feature where collecting a power-up item slows down the buildings. The project already includes a 'Power Up' sprite and a `glideDuration` variable to control the 'Buildings' sprite's speed. Please program the 'Power Up' sprite to generate a clone every few seconds at a random y-position on the right side of the screen that moves toward the left. When the cat touches a 'Power Up' clone, delete the clone and increase the `glideDuration` of the buildings to slow them down. \\
\hline
\vspace{2pt}
\begin{minipage}{0.48\linewidth}
\centering
\includegraphics[width=\linewidth]{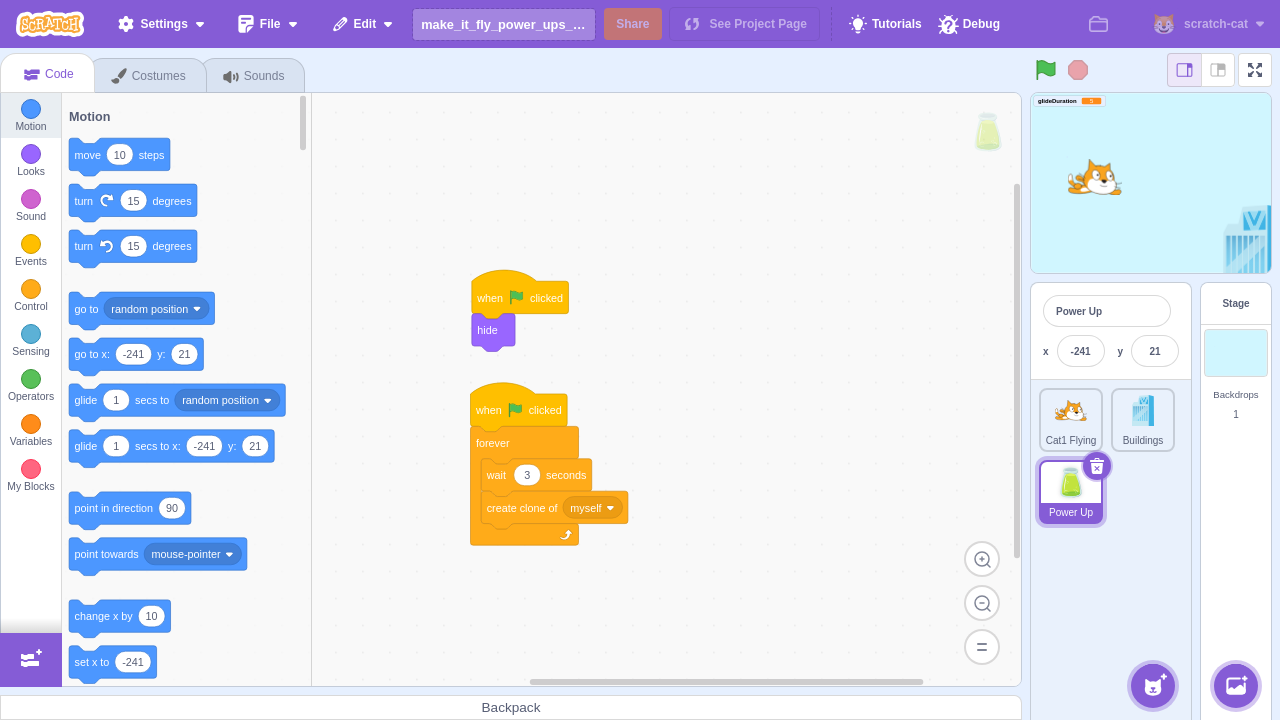} 
\centerline{\small (a) Initial Project State}
\end{minipage}
\hfill
\begin{minipage}{0.48\linewidth}
\centering
\includegraphics[width=\linewidth]{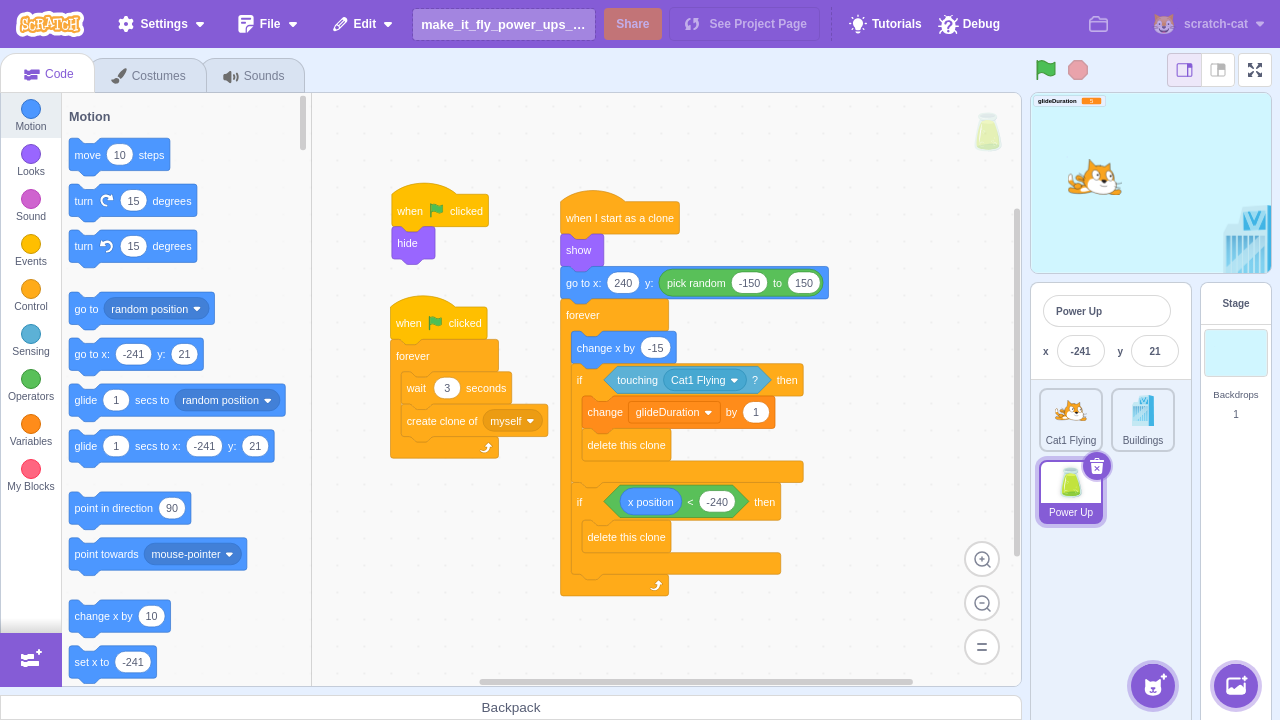}
\centerline{\small (b) Golden Project State}
\end{minipage} 
\vspace{2pt} \\
\hline
\textbf{Test Cases:} 1) 'Power Up' clones spawn on right and move left. 2) Clone deleted/hidden upon touching 'Cat'. 3) Building scroll speed decreases (glide duration increases) after collecting 2 items. 4) Periodic spawn (at least 2 clones in set time). 5) Random spawn Y position. \\
\hline
\hline
\textbf{Type:} Extend \\
\textbf{Instruction:} This is a basic maze navigation game where players control a ball sprite using arrow keys to move through a maze. The ball bounces off blue walls and the goal is to reach a target sprite that displays 'You win!' when touched. Create a clone-based trail effect. In the Ball sprite, periodically create clones while moving. In 'when I start as a clone', set the clone's ghost effect to 0, then in a repeat loop increase ghost by a small amount until the clone becomes invisible, then delete the clone. Ensure clones do not affect gameplay by spacing clone creation. \\
\hline
\vspace{2pt}
\begin{minipage}{0.48\linewidth}
\centering
\includegraphics[width=\linewidth]{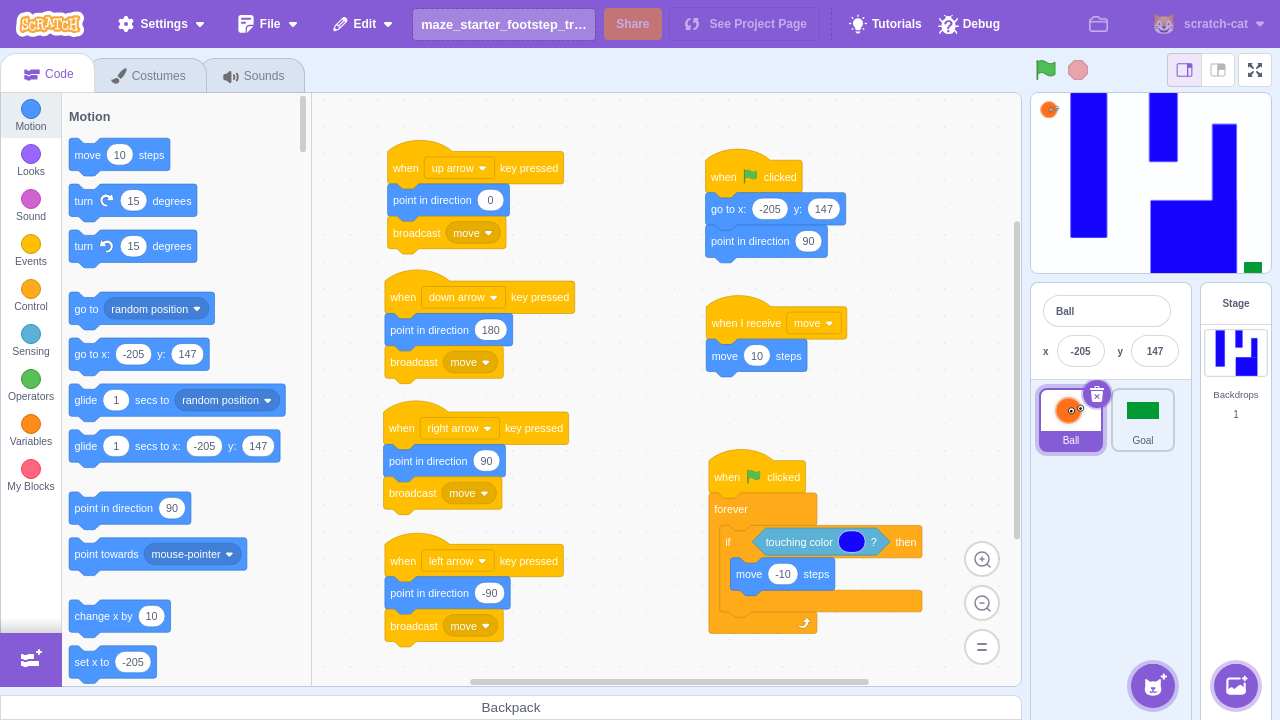} 
\centerline{\small (a) Initial Project State}
\end{minipage}
\hfill
\begin{minipage}{0.48\linewidth}
\centering
\includegraphics[width=\linewidth]{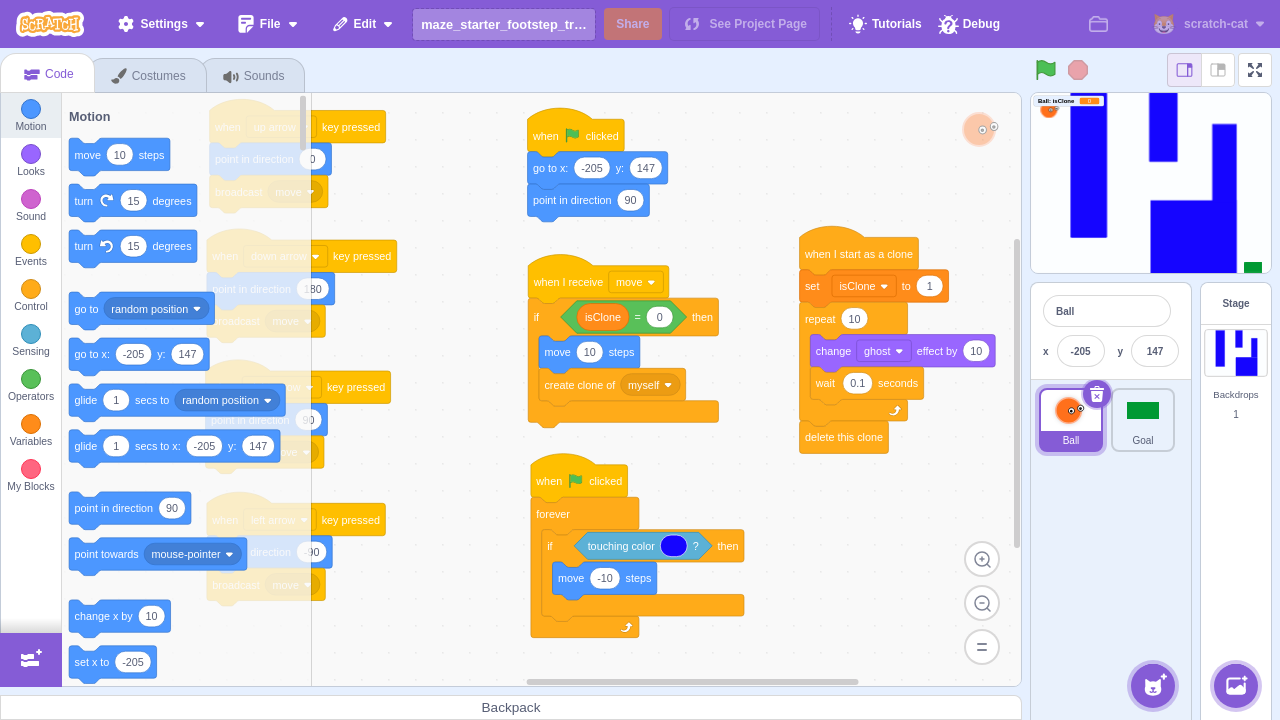}
\centerline{\small (b) Golden Project State}
\end{minipage} 
\vspace{2pt} \\
\hline
\textbf{Test Cases:} 1) Clones created while Ball moves. 2) Ghost effect increases over time (fade out). 3) Clones self-delete (at least 1 deleted). 4) Active clone count $\le$ 30 to prevent overflow. \\
\hline

\hline
\multicolumn{1}{|c|}{\textbf{Category 4: Compute Tasks}} \\
\hline
\textbf{Type:} Compute \\
\textbf{Instruction:} Complete the given Scratch starter project that asks the user to input 5 numbers one by one, adds them to a list, sorts the list in ascending order, and then says the sorted numbers joined by spaces. \\
\hline
\vspace{2pt}
\begin{minipage}{0.48\linewidth}
\centering
\includegraphics[width=\linewidth]{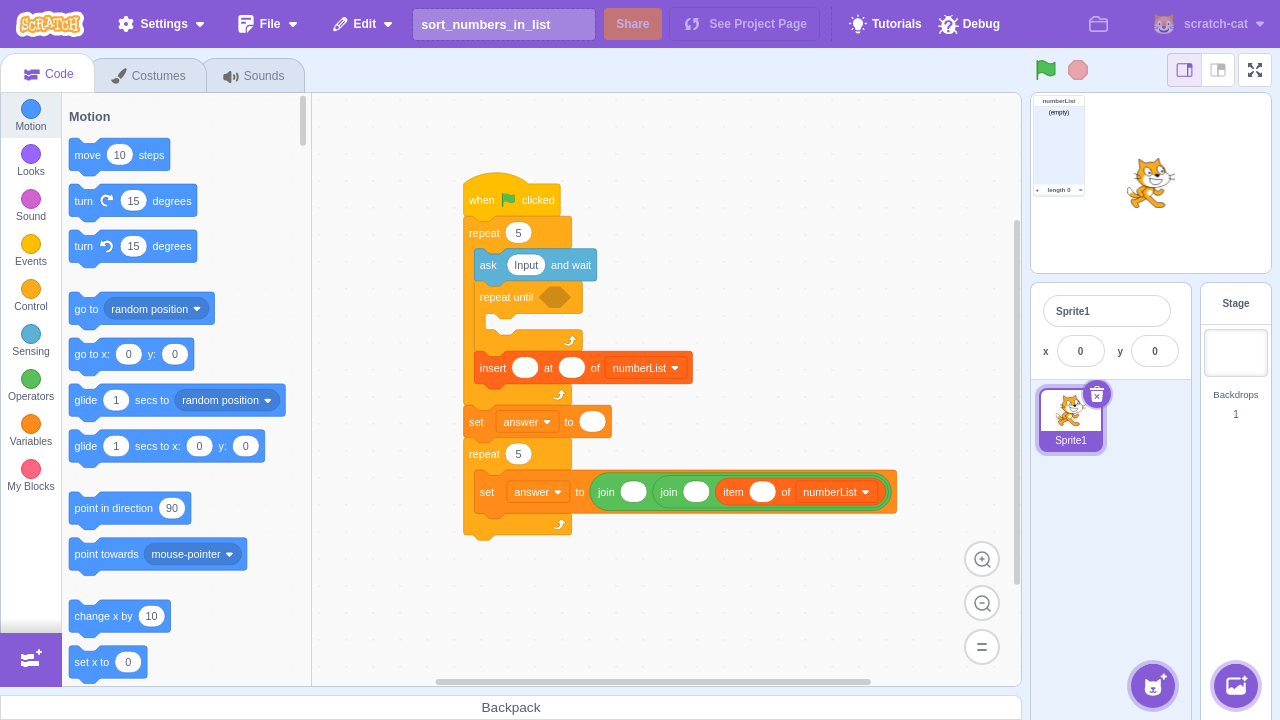} 
\centerline{\small (a) Initial Project State}
\end{minipage}
\hfill
\begin{minipage}{0.48\linewidth}
\centering
\includegraphics[width=\linewidth]{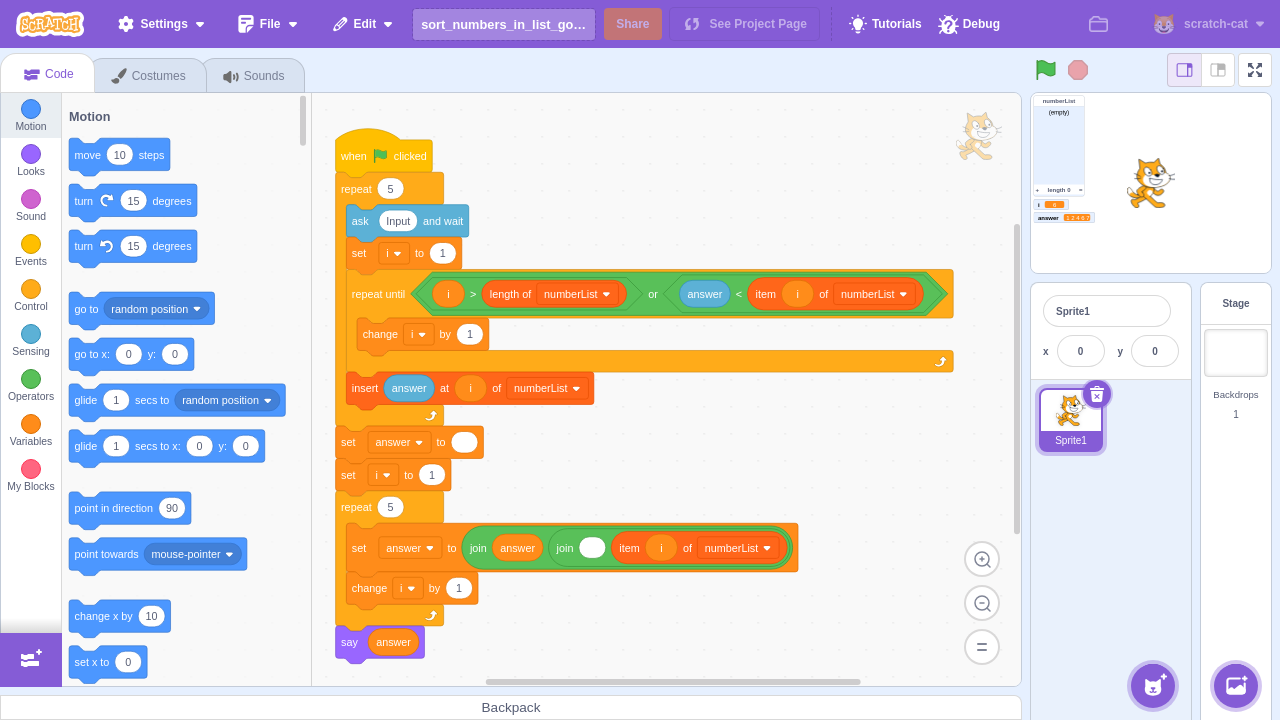}
\centerline{\small (b) Golden Project State}
\end{minipage} 
\vspace{2pt} \\
\hline
\textbf{Test Cases:} 1) Input: [5, 1, 4, 2, 8] $\rightarrow$ Output: '1 2 4 5 8'. 2) Input: [1, 2, 3, 4, 5] $\rightarrow$ Output: '1 2 3 4 5'. 3) Input: [5, 4, 3, 2, 1] $\rightarrow$ Output: '1 2 3 4 5'. \\
\hline
\hline
\textbf{Type:} Compute \\
\textbf{Instruction:} Complete the given Scratch starter project that takes a string as input and checks if it is a palindrome. The sprite should say 'True' if it is, and 'False' otherwise. \\
\hline
\vspace{2pt}
\begin{minipage}{0.48\linewidth}
\centering
\includegraphics[width=\linewidth]{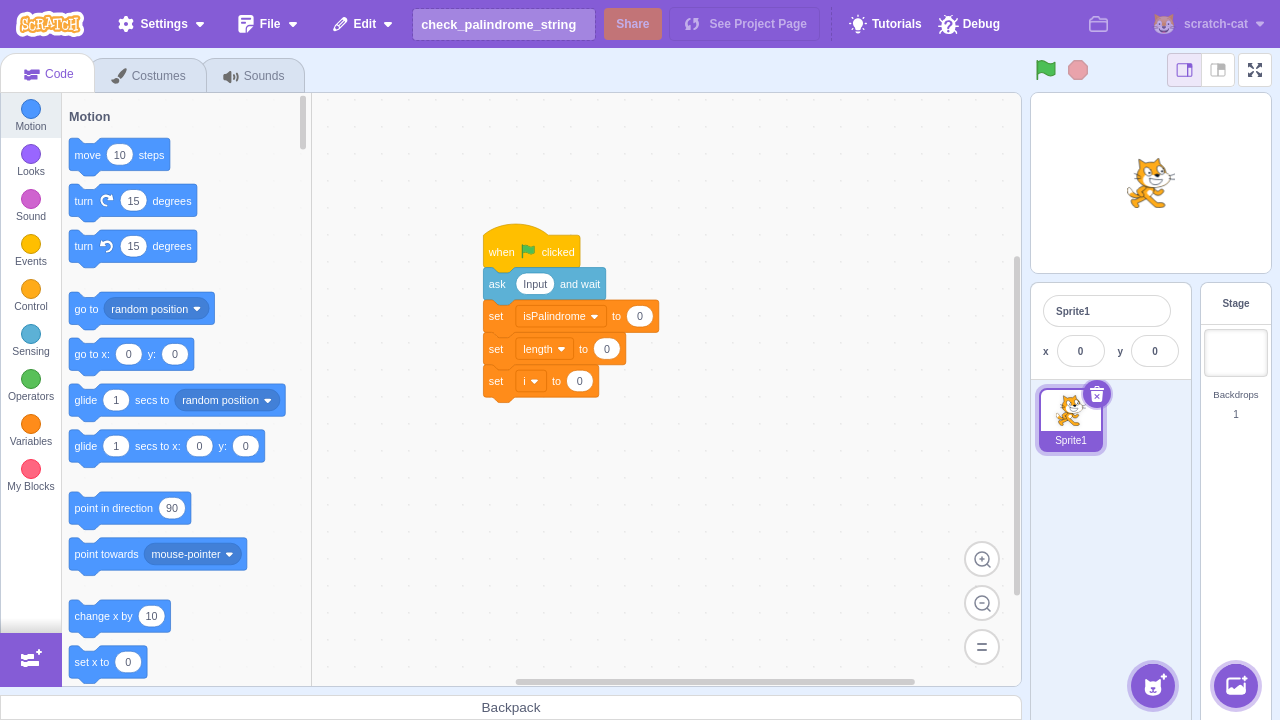} 
\centerline{\small (a) Initial Project State}
\end{minipage}
\hfill
\begin{minipage}{0.48\linewidth}
\centering
\includegraphics[width=\linewidth]{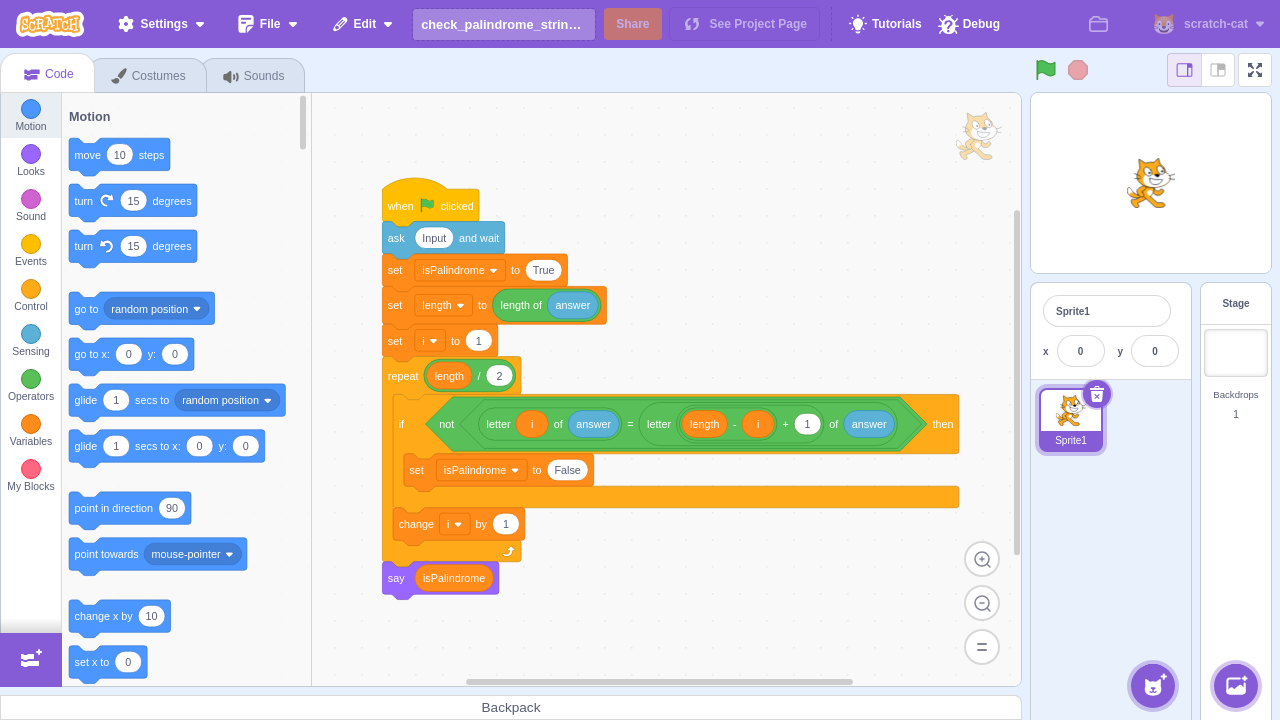}
\centerline{\small (b) Golden Project State}
\end{minipage} 
\vspace{2pt} \\
\hline
\textbf{Test Cases:} 1) Input: 'radar' $\rightarrow$ Output: 'True'. 2) Input: 'hello' $\rightarrow$ Output: 'False'. 3) Input: '12321' $\rightarrow$ Output: 'True'. \\
\hline

\end{longtable}

\clearpage

\section{Detailed Action Space of Primitive Mode}
\label{sec:primitive_action_space}

This section provides the complete specification of the primitive action space $\mathcal{A}_{\text{prim}}$ used in \bench. The action space consists of low-level GUI primitives that mirror human operations, including mouse interactions (click, drag-and-drop, scroll) and keyboard operations (type, key, hotkey). Each action supports flexible parameter formats to accommodate both coordinate-based and index-based targeting strategies. Figure~\ref{fig:primitive_schema} presents the formal JSON schema defining all available actions, and their required parameters.

\begin{figure}[htb!]
\centering
\begin{minipage}{\linewidth}
\begin{lstlisting}
{
  "envelope": {"api": "string", "args": "object"},
  "actions": [
    {"name":"click","args":{"oneOf":[{"x":"number","y":"number"},{"index":"int>=0"}],"button?":"enum(left,right,middle)"}},
    {"name":"double_click","args":{"oneOf":[{"x":"number","y":"number"},{"index":"int>=0"}],"button?":"enum(left,right,middle)"}},
    {"name":"drag_and_drop","args":{"oneOf":[{"start_x":"number","start_y":"number","end_x":"number","end_y":"number"},{"start_index":"int>=0","end_index":"int>=0"},{"start_x":"number","start_y":"number","end_index":"int>=0"},{"start_index":"int>=0","end_x":"number","end_y":"number"}],"duration?":"number>=0"}},
    {"name":"scroll","args":{"direction":"enum(up,down,left,right)","amount?":"number>=1","at?":{"oneOf":[{"x":"number","y":"number"},{"index":"int>=0"},"null"]}}},
    {"name":"type","args":{"text":"string"}},
    {"name":"key","args":{"key":"string"}},
    {"name":"hotkey","args":{"keys":"string[2..]"}},
    {"name":"done","args":{}},
    {"name":"failed","args":{}}
  ]
}

\end{lstlisting}
\end{minipage}
\caption{JSON schema of the primitive action space $\mathcal{A}_{\text{prim}}$ in \bench. This specification defines low-level GUI primitives that agents must execute in primitive mode. The action space includes: mouse operations (\texttt{click}, \texttt{double\_click}, \texttt{drag\_and\_drop}, \texttt{scroll}) with coordinate-based or index-based targeting; keyboard operations (\texttt{type}, \texttt{key}, \texttt{hotkey}); and task completion signals (\texttt{done}, \texttt{failed}).}
\label{fig:primitive_schema}
\end{figure}

\section{Detailed Action Space of Composite Mode}
\label{sec:composite_action_space}

This section provides the complete specification of the composite action space $\mathcal{A}_{\text{comp}}$ used in \bench. Unlike primitive mode, composite mode exposes high-level semantic APIs that abstract away GUI coordinates and drag-and-drop mechanics. Agents manipulate Scratch programs through operations such as adding blocks, connecting blocks with specific placement strategies, setting block field values, and deleting blocks. This abstraction isolates program logic from visuomotor control, enabling pure evaluation of reasoning capabilities. Figure~\ref{fig:composite_schema} presents the formal JSON schema defining all available APIs, and their parameters.

\begin{figure}[htb!]
\centering
\begin{minipage}{\linewidth}
\begin{lstlisting}
{
  "envelope": {"api": "string", "args": "object"},
  "apis": [
    {"name":"select_sprite","args":{"name":"string"}},
    {"name":"select_stage","args":{}},
    {"name":"add_block","args":{"blockType":"string","creation?":{"variableName?":"string","listName?":"string"}}},
    {"name":"connect_blocks","args":{"sourceBlockIndex":"int","targetBlockIndex":"int","placement":{"kind":"enum","inputName?":"string"}}},
    {"name":"set_block_field","args":{"blockIndex":"int","fieldName":"string","value":"any"}},
    {"name":"delete_block","args":{"blockIndex":"int"}},
    {"name":"done","args":{}},
    {"name":"failed","args":{}}
  ]
}

\end{lstlisting}
\end{minipage}
\caption{JSON schema of the composite action space $\mathcal{A}_{\text{comp}}$ in \bench. This specification defines high-level semantic APIs that agents use in composite mode to manipulate Scratch programs without visual grounding. The action space includes: target selection (\texttt{select\_sprite}, \texttt{select\_stage}), block manipulation (\texttt{add\_block} with optional variable/list creation, \texttt{connect\_blocks} with placement strategies, \texttt{set\_block\_field} for parameter configuration, \texttt{delete\_block}), and task completion signals.}
\label{fig:composite_schema}
\end{figure}

\section{Example of Composite Mode Observation}
\label{sec:comp_observations}

This section illustrates the observation format provided to agents in composite mode. As described in Section~\ref{sec:modes}, composite mode replaces visual screenshots with structured textual representations that capture the current state of the Scratch program. The observation includes environmental context (available variables, lists, and editing targets) and a pseudocode representation of the block structure with indexed blocks for manipulation. Figure~\ref{fig:composite_obs} shows a concrete example demonstrating how a simple Scratch program (a sprite moving with rotation style control) is represented in this textual format.

\begin{figure}[htb!]
  \centering
  \vspace{0.3em}
  \begin{minipage}{\linewidth}
  \begin{lstlisting}
## Current Editing Target
Sprite1

## Target Variables In Scope
name: my variable, scope: all

## Target Lists In Scope
None

## All Available Targets
Stage, Sprite1

## Blocks Pseudocode
#1 [top] event_whenflagclicked
#2 motion_setrotationstyle
- field STYLE: "left-right" {choices: {"styleOptions":["left-right","don't rotate","all around"]}}
#3 control_forever
- SUBSTACK:
  #4 motion_movesteps
  - input STEPS: 10 (math_number)
  #5 motion_ifonedgebounce
  \end{lstlisting}
  \end{minipage}
  \caption{Example of the composite mode observation. It includes environmental context (variables, lists, and targets) and the block structure (pseudocode).}
  \label{fig:composite_obs}
\end{figure}

\section{Single-Step Drag-and-Drop Task Examples}
\label{app:drag_examples}

Table \ref{tab:drag_examples} presents representative examples from the Single-Step Drag Benchmark used in RQ2. Each row shows one atomic drag-and-drop task with its type, instruction, and the corresponding Scratch canvas screenshot.

\begin{table}[htb!]
\centering
\caption{Representative examples from the Single-Step Drag Benchmark used in RQ2 to isolate drag-and-drop precision. The benchmark contains 60 atomic tasks spanning two interaction types: \textbf{Direct Connection} (vertically stacking command blocks by aligning top/bottom connection points) and \textbf{Slot Insertion} (inserting reporter/boolean blocks into input slots within command blocks). Each row shows one task with its type, natural language instruction describing the required drag operation, and the corresponding Scratch canvas screenshot showing the initial block arrangement.}
\label{tab:drag_examples}
\small
\setlength{\tabcolsep}{6pt}
\begin{tabular}{>{\centering\arraybackslash}m{2.5cm}>{\centering\arraybackslash}m{5.5cm}>{\centering\arraybackslash}m{7cm}}
\toprule
\textbf{Type} & \textbf{Instruction} & \textbf{Screenshot} \\
\midrule
Direct Connection & 
\textit{Drag the 'when green flag clicked' block and place it directly above the 'set rotation style' block to connect them.} & 
\includegraphics[width=6.5cm]{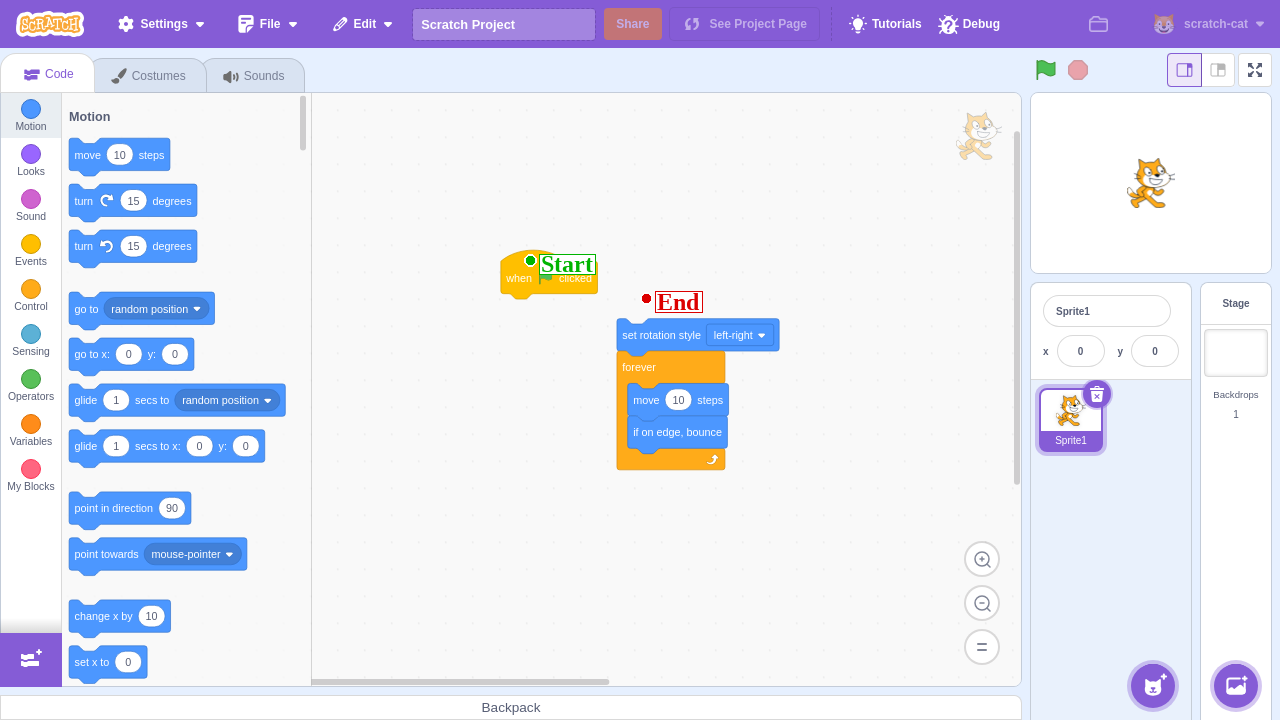} \\
\hline
Direct Connection & 
\textit{Drag the stack starting with the 'if then' block and insert it between the first 'if then' block and the second 'if then' block inside the 'forever' block.} & 
\includegraphics[width=6.5cm]{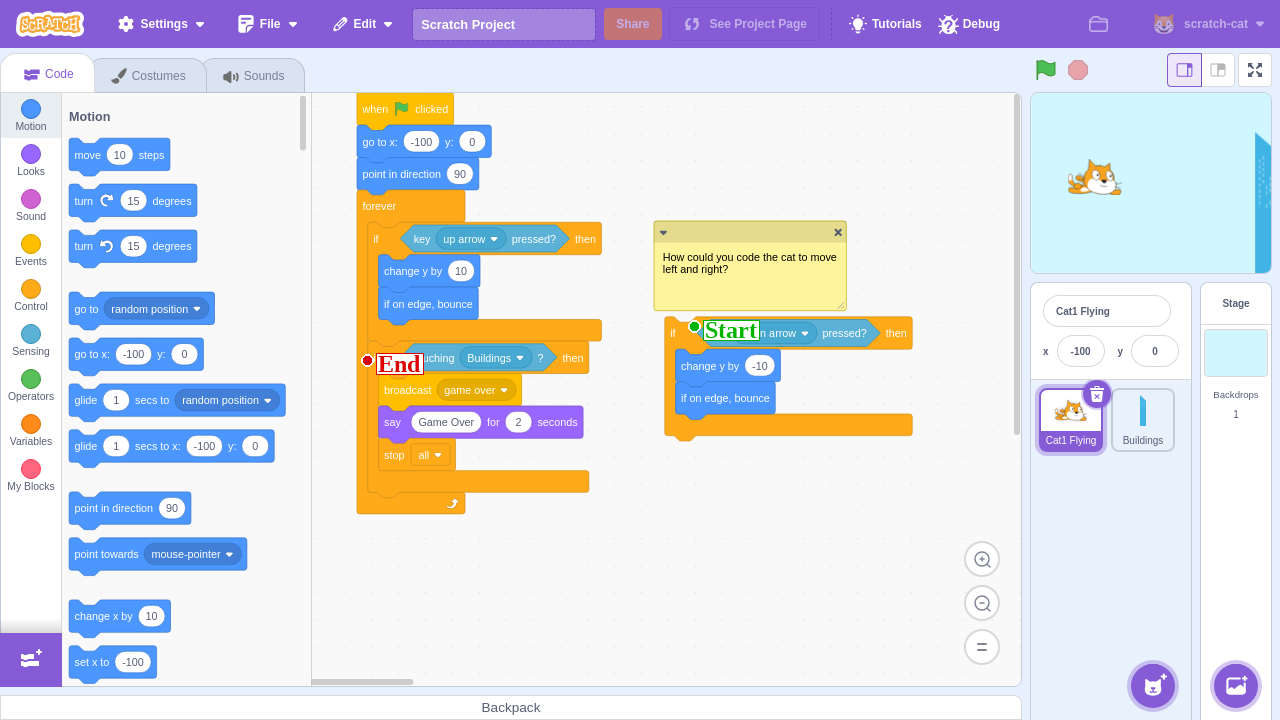} \\
\hline
Slot Insertion & 
\textit{Drag the 'say for 2 seconds' block and connect it into the substack of the 'if then' block.} & 
\includegraphics[width=6.5cm]{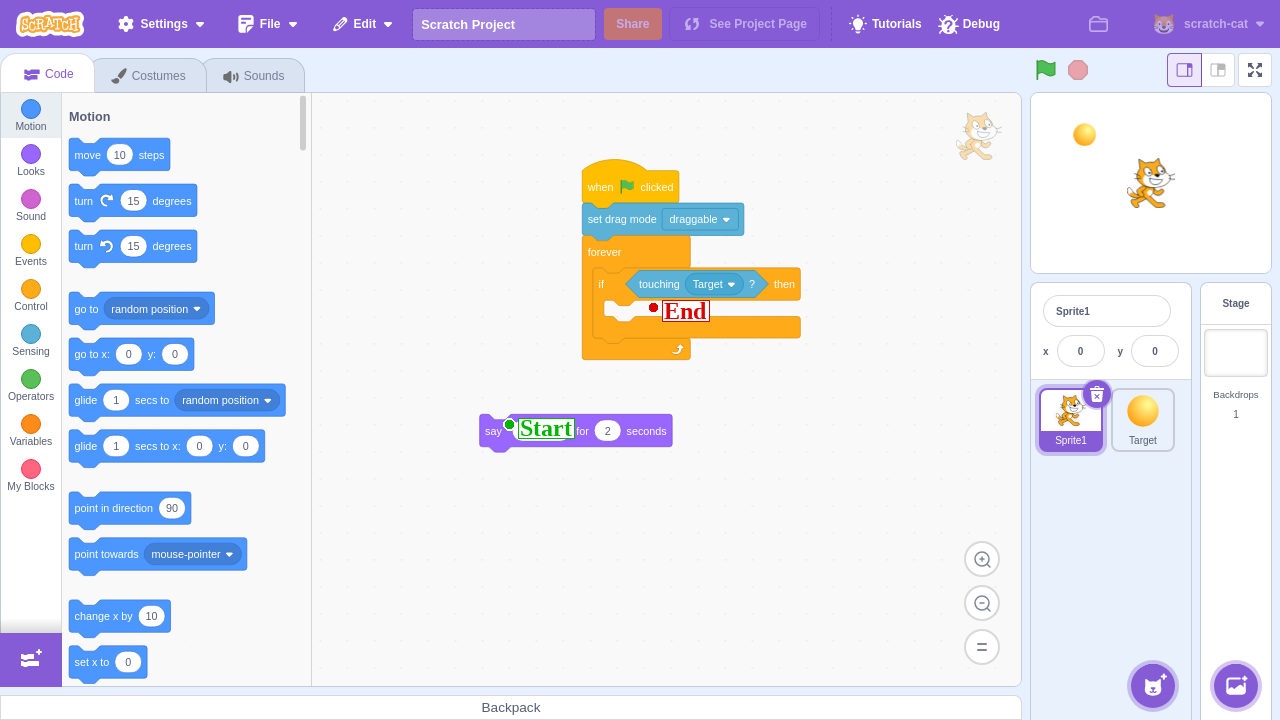} \\
\hline
Slot Insertion & 
\textit{Drag the 'glideTime' data variable block from the flyout and connect it into the first value input slot of the 'glide secs to xy' block.} & 
\includegraphics[width=6.5cm]{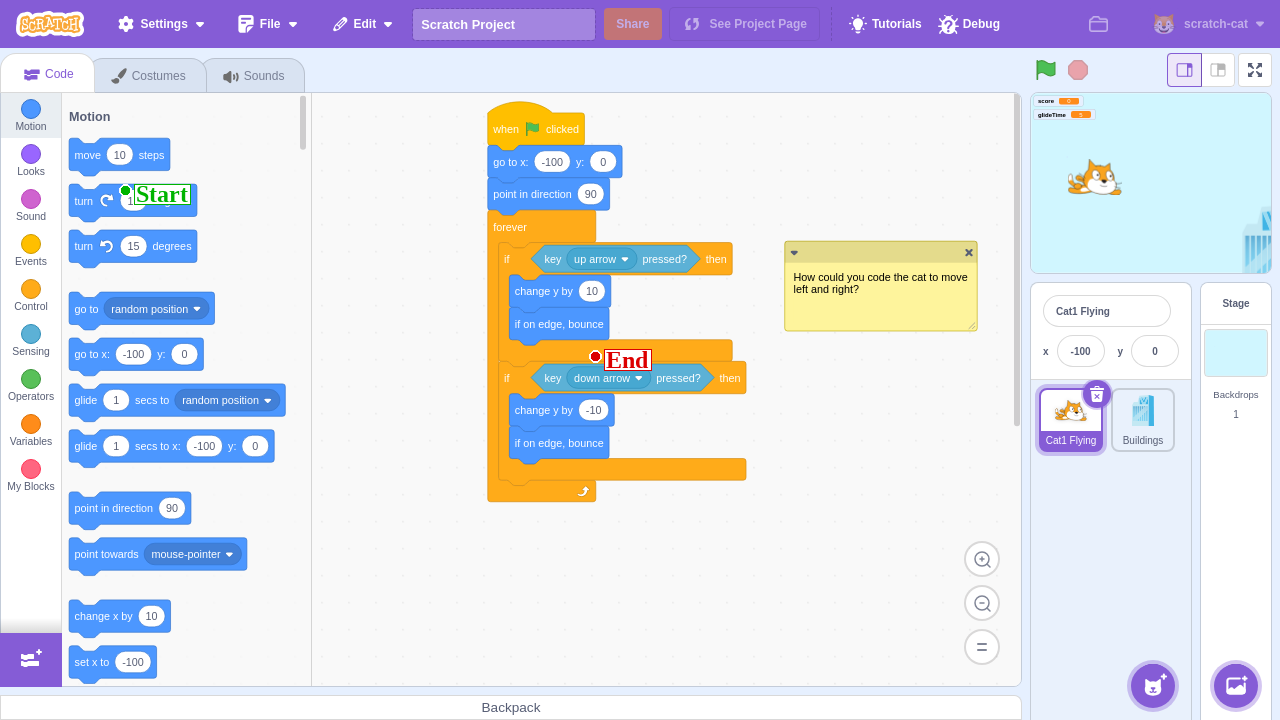} \\
\bottomrule
\end{tabular}
\end{table}

\clearpage

\section{Visual Perception Task Examples}
\label{app:perception_examples}

Table \ref{tab:perception_examples} presents representative examples from the static visual perception benchmark used in RQ3. Each row illustrates one of the three evaluation dimensions with the corresponding question, screenshot, and ground-truth answer.

\begin{table}[htb!]
\centering
\caption{Representative examples from the static Visual Perception QA Benchmark used in RQ3 to diagnose perception deficits. The benchmark contains 200 manually curated samples targeting three recurring error patterns observed in failure trajectories: \textbf{Connection} (false positives in detecting block connections), \textbf{Existence} (false negatives when blocks are occluded or partially visible), and \textbf{Field Value} (reading errors for dropdown menus and numeric inputs, evaluated with both equivalence checking and exact string matching). Each row shows one example with its evaluation dimension, the question posed to the model, the Scratch canvas screenshot, and the ground-truth answer (YES/NO for binary questions, exact text for field values).}
\label{tab:perception_examples}
\small
\setlength{\tabcolsep}{6pt}
\begin{tabular}{>{\centering\arraybackslash}m{2.2cm}>{\centering\arraybackslash}m{4.8cm}>{\centering\arraybackslash}m{6cm}>{\centering\arraybackslash}m{1.5cm}}
\toprule
\textbf{Type} & \textbf{Question} & \textbf{Screenshot} & \textbf{Answer} \\
\midrule
Connection & 
\textit{Is the bottommost `if then` block directly connected to the block directly above it?} & 
\includegraphics[width=5.5cm]{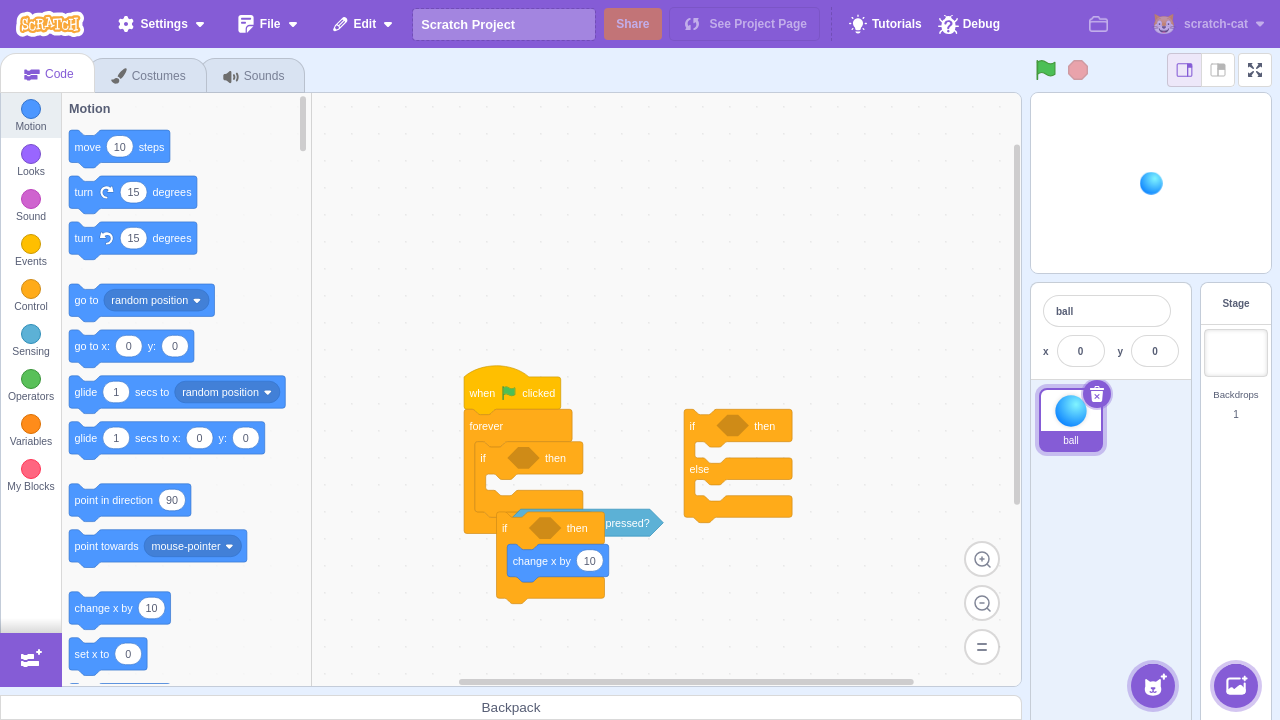} & 
\textit{NO} \\
\hline
Existence & 
\textit{Is there any `if then else` block that is visible in the canvas?} & 
\includegraphics[width=5.5cm]{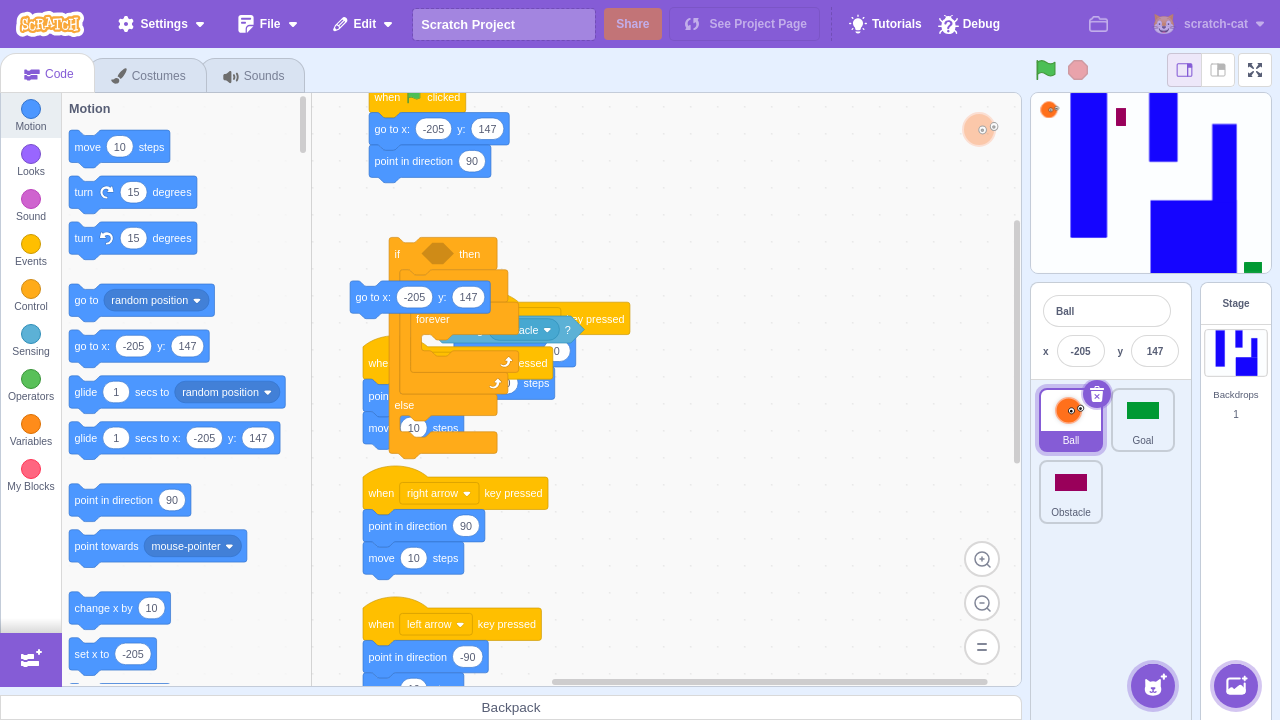} & 
\textit{YES} \\
\hline
Field Value (Equivalence)& 
\textit{Is the dropdown field in the `key [key] pressed? ` block set to `space`} & 
\includegraphics[width=5.5cm]{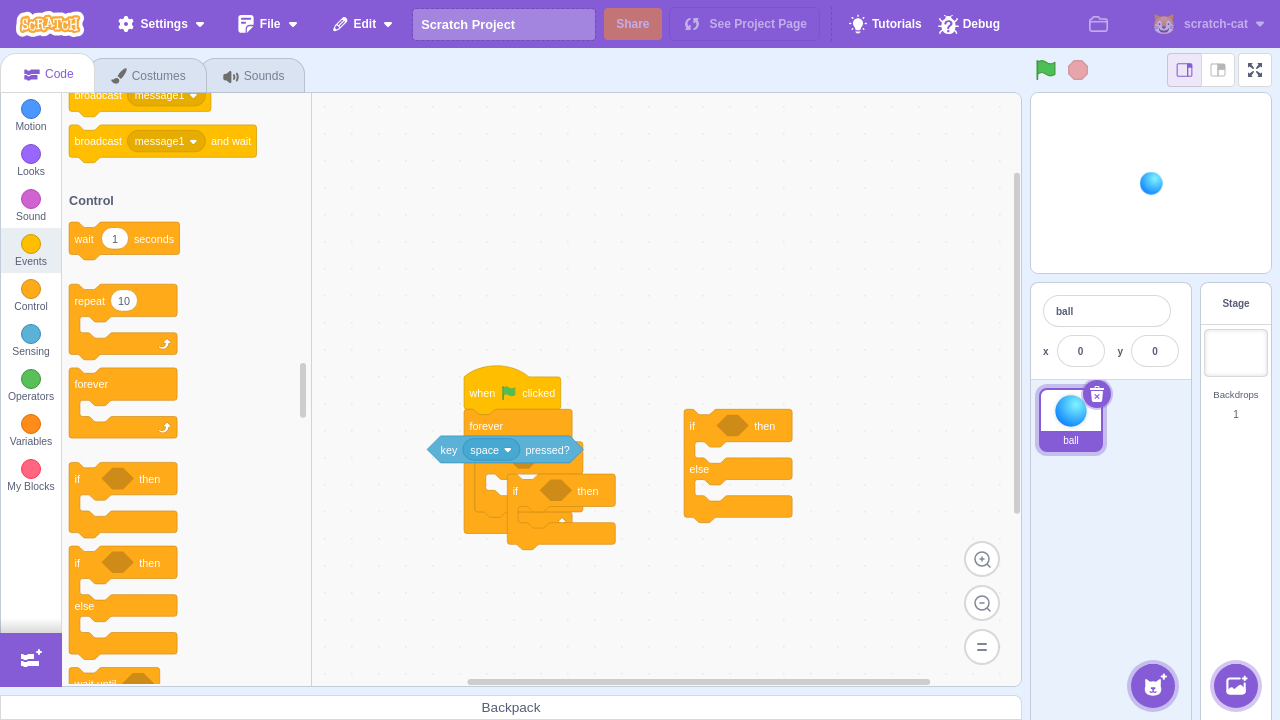} & 
\textit{YES} \\
\hline
Field Value (Exact Match)& 
\textit{What number is shown in the left numeric field of the rightmost `letter of` block?} & 
\includegraphics[width=5.5cm]{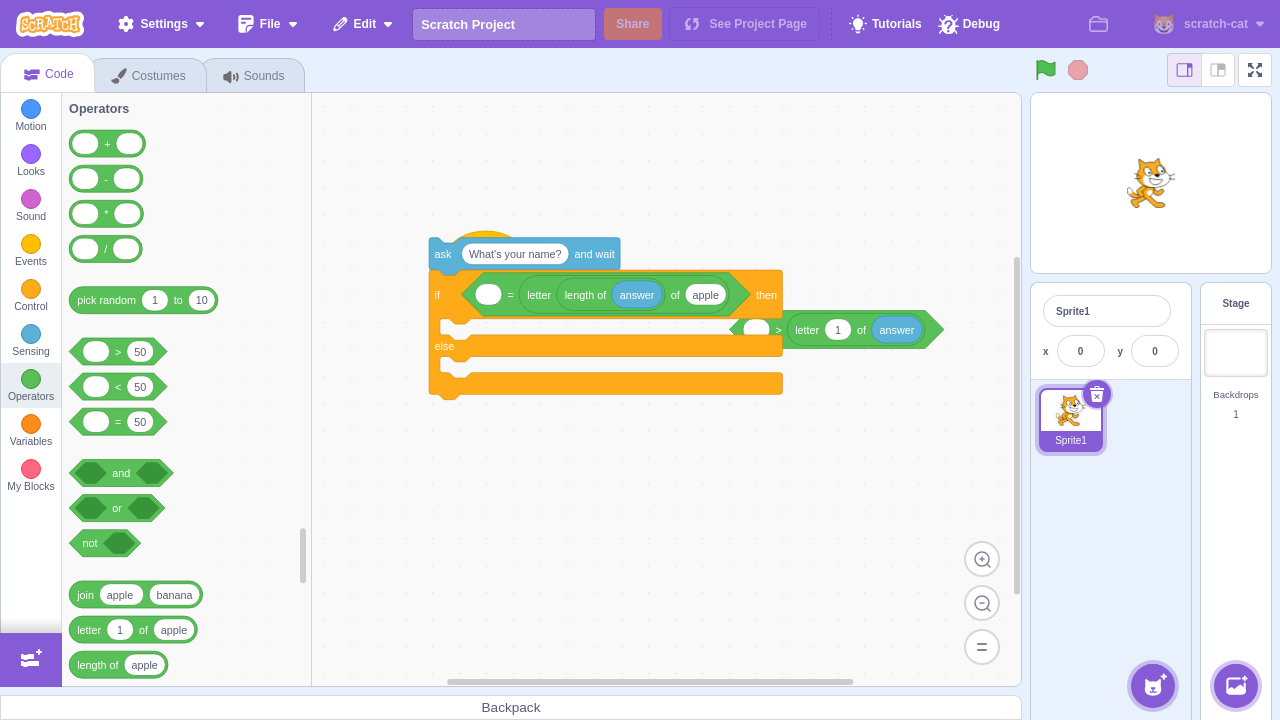} & 
\textit{1} \\
\bottomrule
\end{tabular}
\end{table}

\clearpage

\section{Results for Single-Step Drag Benchmark in Task Categories}
\label{app:drag_category_results}

This section provides a detailed breakdown of the Single-Step Drag Benchmark results (reported in aggregate in Table~\ref{tab:rq2_results_all}) by interaction type. We distinguish between two fundamental drag-and-drop scenarios in Scratch: \textbf{Direct Connection} (T1), which involves vertically stacking command blocks by aligning their connection points, and \textbf{Slot Insertion} (T2), which involves inserting reporter or boolean blocks into specific input slots within command blocks. This decomposition reveals that models exhibit different performance patterns across interaction types, with slot insertion generally showing higher endpoint component accuracy but lower start-point accuracy compared to direct connection. Table~\ref{tab:rq2_results_t1_t2} presents the complete results.

\begin{table}[htb!]
    \centering
    \caption{Detailed breakdown of Single-Step Drag Benchmark results by interaction type. \textbf{T1 (Direct Connection)} involves vertically stacking command blocks by connecting their top/bottom edges. \textbf{T2 (Slot Insertion)} involves inserting reporter or boolean blocks into specific input slots within command blocks.}
    \label{tab:rq2_results_t1_t2}
    \resizebox{\textwidth}{!}{%
    \begin{tabular}{ll|ccccccc|ccccccc}
        \toprule
        & & \multicolumn{7}{c|}{T1 (Direct Connection)} & \multicolumn{7}{c}{T2 (Slot Insertion)} \\
        \cmidrule(lr){3-9} \cmidrule(lr){10-16}
        & & \multicolumn{3}{c}{Success Rate (\%)} & \multicolumn{2}{c}{Comp. Acc. (\%)} & \multicolumn{2}{c|}{Spatial Err. (px)} & \multicolumn{3}{c}{Success Rate (\%)} & \multicolumn{2}{c}{Comp. Acc. (\%)} & \multicolumn{2}{c}{Spatial Err. (px)} \\
        \cmidrule(lr){3-5} \cmidrule(lr){6-7} \cmidrule(lr){8-9} \cmidrule(lr){10-12} \cmidrule(lr){13-14} \cmidrule(lr){15-16}
        Model & Setting & @1 & @2 & @3 & Start & End & Start & End & @1 & @2 & @3 & Start & End & Start & End \\
        \midrule
        \multirow{3}{*}{GPT-5} 
          & Baseline  & \rqTwoSR{23.33}{23.33} & \rqTwoSR{26.67}{26.67} & \rqTwoSR{33.33}{33.33} & \rqTwoAcc{74.44}{74.44} & \rqTwoAcc{22.39}{22.39} & \rqTwoErr{18.50}{18.50} & \rqTwoErr{33.63}{33.63} & \rqTwoSR{23.33}{23.33} & \rqTwoSR{30.00}{30.00} & \rqTwoSR{36.67}{36.67} & \rqTwoAcc{58.89}{58.89} & \rqTwoAcc{45.28}{45.28} & \rqTwoErr{34.42}{34.42} & \rqTwoErr{40.40}{40.40} \\
          & GT Start  & \rqTwoSR{6.67}{6.67} & \rqTwoSR{13.33}{13.33} & \rqTwoSR{20.00}{20.00} & \rqTwoAcc{100.00}{100.00} & \rqTwoAcc{10.00}{10.00} & \textemdash & \rqTwoErr{40.61}{40.61} & \rqTwoSR{46.67}{46.67} & \rqTwoSR{60.00}{60.00} & \rqTwoSR{73.33}{73.33} & \rqTwoAcc{98.89}{98.89} & \rqTwoAcc{50.56}{50.56} & \rqTwoErr{60.00}{60.00} & \rqTwoErr{36.20}{36.20} \\
          & Knowledge & \rqTwoSR{36.67}{36.67} & \rqTwoSR{40.00}{40.00} & \rqTwoSR{50.00}{50.00} & \rqTwoAcc{74.44}{74.44} & \rqTwoAcc{38.81}{38.81} & \rqTwoErr{9.53}{9.53} & \rqTwoErr{28.25}{28.25} & \rqTwoSR{26.67}{26.67} & \rqTwoSR{43.33}{43.33} & \rqTwoSR{53.33}{53.33} & \rqTwoAcc{52.22}{52.22} & \rqTwoAcc{51.06}{51.06} & \rqTwoErr{14.88}{14.88} & \rqTwoErr{23.55}{23.55} \\
        \midrule
        \multirow{3}{*}{Qwen3-VL-32B-Instruct} 
          & Baseline  & \rqTwoSR{20.00}{20.00} & \rqTwoSR{20.00}{20.00} & \rqTwoSR{23.33}{23.33} & \rqTwoAcc{73.33}{73.33} & \rqTwoAcc{22.73}{22.73} & \rqTwoErr{25.72}{25.72} & \rqTwoErr{44.41}{44.41} & \rqTwoSR{33.33}{33.33} & \rqTwoSR{43.33}{43.33} & \rqTwoSR{43.33}{43.33} & \rqTwoAcc{62.22}{62.22} & \rqTwoAcc{53.57}{53.57} & \rqTwoErr{225.43}{225.43} & \rqTwoErr{48.18}{48.18} \\
          & GT Start  & \rqTwoSR{16.67}{16.67} & \rqTwoSR{16.67}{16.67} & \rqTwoSR{16.67}{16.67} & \rqTwoAcc{100.00}{100.00} & \rqTwoAcc{12.22}{12.22} & \textemdash & \rqTwoErr{51.01}{51.01} & \rqTwoSR{53.33}{53.33} & \rqTwoSR{60.00}{60.00} & \rqTwoSR{63.33}{63.33} & \rqTwoAcc{100.00}{100.00} & \rqTwoAcc{52.22}{52.22} & \textemdash & \rqTwoErr{30.62}{30.62} \\
          & Knowledge & \rqTwoSR{20.00}{20.00} & \rqTwoSR{33.33}{33.33} & \rqTwoSR{43.33}{43.33} & \rqTwoAcc{68.89}{68.89} & \rqTwoAcc{32.26}{32.26} & \rqTwoErr{23.29}{23.29} & \rqTwoErr{43.80}{43.80} & \rqTwoSR{36.67}{36.67} & \rqTwoSR{46.67}{46.67} & \rqTwoSR{60.00}{60.00} & \rqTwoAcc{66.67}{66.67} & \rqTwoAcc{51.67}{51.67} & \rqTwoErr{124.71}{124.71} & \rqTwoErr{20.95}{20.95} \\
        \midrule
        \multirow{3}{*}{UI-TARS-1.5-7B} 
          & Baseline  & \rqTwoSR{3.33}{3.33} & \rqTwoSR{6.67}{6.67} & \rqTwoSR{6.67}{6.67} & \rqTwoAcc{61.11}{61.11} & \rqTwoAcc{3.64}{3.64} & \rqTwoErr{150.50}{150.50} & \rqTwoErr{116.46}{116.46} & \rqTwoSR{0.00}{0.00} & \rqTwoSR{6.67}{6.67} & \rqTwoSR{6.67}{6.67} & \rqTwoAcc{37.78}{37.78} & \rqTwoAcc{8.82}{8.82} & \rqTwoErr{293.49}{293.49} & \rqTwoErr{30.07}{30.07} \\
          & GT Start  & \rqTwoSR{6.67}{6.67} & \rqTwoSR{6.67}{6.67} & \rqTwoSR{6.67}{6.67} & \rqTwoAcc{70.00}{70.00} & \rqTwoAcc{9.52}{9.52} & \rqTwoErr{93.13}{93.13} & \rqTwoErr{86.74}{86.74} & \rqTwoSR{20.00}{20.00} & \rqTwoSR{20.00}{20.00} & \rqTwoSR{20.00}{20.00} & \rqTwoAcc{43.33}{43.33} & \rqTwoAcc{46.15}{46.15} & \rqTwoErr{329.16}{329.16} & \rqTwoErr{31.59}{31.59} \\
          & Knowledge & \rqTwoSR{10.00}{10.00} & \rqTwoSR{10.00}{10.00} & \rqTwoSR{13.33}{13.33} & \rqTwoAcc{52.22}{52.22} & \rqTwoAcc{19.15}{19.15} & \rqTwoErr{226.48}{226.48} & \rqTwoErr{109.07}{109.07} & \rqTwoSR{23.33}{23.33} & \rqTwoSR{23.33}{23.33} & \rqTwoSR{23.33}{23.33} & \rqTwoAcc{37.78}{37.78} & \rqTwoAcc{61.76}{61.76} & \rqTwoErr{269.22}{269.22} & \rqTwoErr{35.59}{35.59} \\
        \bottomrule
    \end{tabular}%
    }
\end{table}

\section{BFS-Based Feasible Region Computation}
\label{app:bfs-feasible-regions}

To rigorously quantify localization errors in the Single-Step Drag Benchmark, we developed a BFS-based algorithm to compute the feasible regions for valid start and end positions in each drag-and-drop task. This automated procedure enables precise evaluation of model predictions by identifying all pixel coordinates that yield successful drag operations.

\paragraph{Computing Feasible Start Positions.}
For each task, we first retrieve the opcode and unique identifier of the block to be dragged, which allows us to determine its precise location on the canvas or in the flyout (block palette). 
Using heuristic rules, we compute an initial seed point, typically the geometric center of the block's bounding box. 
We then perform a trial drag operation: for blocks in the flyout, we drag toward the canvas; for blocks already on the canvas, we drag toward a location far from their current position. 
If the block's position changes after the drag, the seed point is marked as feasible. 
If the seed point fails (i.e., the block does not move), we systematically explore neighboring pixels in an expanding radius until a feasible point is identified.

Once a feasible start point is found, we apply BFS to discover all contiguous feasible start positions. 
Starting from the initial feasible point, BFS expands outward by testing each neighboring pixel (4-directional connectivity) and marking it as feasible if a drag operation from that coordinate successfully displaces the block. 
This process continues until no additional feasible neighbors can be found, yielding a complete connected component of valid start coordinates.

\paragraph{Computing Feasible End Positions.}
To determine feasible end positions, we fix a known feasible start point and compute the set of end coordinates that result in successful block connections. 
Using heuristic rules and the identifier of the target block (the block to which we intend to connect), we calculate a seed end point based on the expected connection geometry (e.g., snapping distance, alignment offset). 
We then perform BFS centered at this seed point, testing each candidate end coordinate by executing a full drag-and-drop operation from the fixed start position to the candidate end position.

For each candidate end point, we validate the connection outcome by querying the \texttt{Scratch VM} runtime state. 
Specifically, we designed task-specific verification rules that inspect the parent-child relationships, input slot occupancy, or stack ordering in the block AST to determine whether the drag operation successfully established the intended connection. 
A candidate end point is marked as feasible if and only if the verification rule confirms a correct connection. 
BFS continues to expand until all feasible end coordinates in the neighborhood are identified.

\section{Heuristics Hints for Drag-and-Drop Tasks}
\label{app:heuristics-for-drag-and-drop}

In the \textit{Knowledge} setting of Single-Step Drag Benchmark task, we provide the following heuristic hints (Figure~\ref{fig:coordinate_heuristics}):
\begin{figure}[htb!]
  \centering
  \vspace{0.3em}
  \begin{minipage}{\linewidth}
  \begin{lstlisting}[basicstyle=\ttfamily\footnotesize, breaklines=true]
### Experience & Heuristics for Coordinate Prediction
  
1. Start Point
Target: The visible body of the Source Block (or Top Block if moving a stack).
Heuristic: Set start_point near the Left Edge (e.g., 10-20px from the left).
Note: Grabbing the Top Block prevents splitting the stack; grabbing the left side facilitates alignment.

2. End Point - Vertical Stacking (Command Blocks)
Context: Connecting single blocks or entire stacks vertically.

Appending (Bottom):
  - Align X with Target's Left Edge.
  - Set Y 15-20px below Target's Bottom Edge.

Prepending (Top):
  - Align X with Target's Left Edge.
  - Height Adjustment: Set Y to (Target_Top_Y - Source_Height).
  - Reasoning: Since the cursor holds the top of the source stack, lift it by its total height so its bottom connects with the target's top.

3. End Point - Nesting (Inside C-Blocks)
Context: Placing a block/stack inside container blocks like "Forever".
X: Indent 15-20px to the right of Target's Left Edge.
Y: Position 15-20px below the bottom edge of the C-block's "top arm" (inside the mouth).

4. End Point - Parameter Insertion (Values & Booleans)
Context: Dropping round (Reporter) or hexagonal (Boolean) blocks into input slots.
Target: The specific empty input slot on the Target Block.
Heuristic: Set end_point to the geometric center of that target input slot.
  \end{lstlisting}
  \end{minipage}
  \caption{Heuristics and experience rules provided to the model for coordinate prediction. These rules guide the calculation of precise start and end points for different drag-and-drop scenarios.}
  \label{fig:coordinate_heuristics}
\end{figure}

\section{Project Serialization Example}
\label{sec:serialization_example}

To enable LLMs to process graphical Scratch projects provided in the \texttt{.sb3} format, we serialize them into a structured text format. This representation preserves the project's hierarchical structure (Stage vs. Sprites), metadata (variables, costumes, sounds), and logic scripts (represented as pseudo-code). Figure~\ref{fig:serialization_example} shows an example of a serialized project:

\begin{figure}[htb!]
\centering
\begin{minipage}{\linewidth}
\begin{lstlisting}[basicstyle=\ttfamily\scriptsize, breaklines=true]
## Stage (Stage)
### Scripts: *No scripts found*
### Variables: **my variable**: 0
### Costumes: **backdrop1** (svg), **Xy-grid** (png)
### Sounds: **pop** (wav)

## Sprite1 (Sprite)
### Scripts:
```
when green flag clicked
go to x: 0 y: 0
set size to 50%
broadcast {BROADCAST_OPTION}
```
```
when I receive enter mode
stop other scripts in sprite
[sensing_setdragmode]
ask join join I am at X: [xpos] Y: [ypos]. Change X by? and wait
change x by [sensing_answer]
... (Ask and Change Y logic omitted for brevity)
broadcast {BROADCAST_OPTION}
```
```
when I receive drag mode
[sensing_setdragmode]
forever
  say join join X:  round [motion_xposition] join  Y:  round [motion_yposition]
```

### Variables: *No variables*
### Costumes: **costume1** (svg), **costume2** (svg)
### Sounds: **Meow** (wav)

## Button (Sprite)
### Scripts:
```
when this sprite clicked
broadcast {BROADCAST_OPTION}
hide
```
```
when green flag clicked
go to x: -187 y: -150
```
... (Other scripts omitted)
\end{lstlisting}
\end{minipage}
\caption{An example of the serialized textual representation of a Scratch project used in \bench task construction.}
\label{fig:serialization_example}
\end{figure}

\section{Example of Element List Observation}
\label{sec:element_list_example}

This section illustrates the structured element list $\mathcal{E}$ provided to agents in primitive mode observation space. As described in Section~\ref{sec:modes}, to assist visual grounding and mitigate pixel-level ambiguity, we augment raw screenshots with an indexed textual enumeration of interactive UI elements. This hybrid representation combines visual information with structured metadata, enabling agents to reference elements either by pixel coordinates or by index. The element list is constructed by merging DOM elements with OCR-detected text regions (see Appendix~\ref{app:element-list-fusion} for the fusion algorithm). Figure~\ref{fig:element_list_example} shows a concrete example of the element list format.

\begin{figure}[htb!]
\centering
\begin{minipage}{\linewidth}
\begin{lstlisting}[basicstyle=\ttfamily\scriptsize, breaklines=true]
index    type             text                                                  position
0        inputs           value: Scratch Project placeholder: Project title...  (412, 8) 184x33
1        inputs           value: Sprite1 placeholder: Name                      (1043, 295) 128x32
3        inputs           value: 0 placeholder: y                               (1131, 335) 48x32
4        sprites          Sprite1 duplicate export delete                       (1039, 388) 64x64
6        blocks           event_whenflagclicked on canvas                       (310, 81) 99x52
7        blocks           motion_gotoxy on canvas                               (310, 124) 128x40
...
33       blocks           motion_movesteps                                      (68, 137) 104x40
43       blocks           motion_pointtowards                                   (68, 536) 180x40
44       category_menu    Motion                                                (1, 93) 60x45
45       category_menu    Looks                                                 (1, 138) 60x45
...
53       text             ScRATCH                                               (15, 10) 70x26
54       text             Settings                                              (136, 12) 56x21
55       text             File                                                  (256, 14) 28x17
56       text             Edit                                                  (351, 15) 29x16
...
\end{lstlisting}
\end{minipage}
\caption{An example of the structured element list $\mathcal{E}$ provided to agents in primitive mode observation space.}
\label{fig:element_list_example}
\end{figure}

\section{Example of Evaluation Script}
\label{sec:eval_script_example}

Figure~\ref{fig:eval_script_code} shows a simplified example of the JavaScript evaluation logic injected into the browser. It demonstrates how we interface with the \texttt{Scratch VM} to trigger events (e.g., Green Flag) and verify internal states (e.g., sprite coordinates).

\begin{figure}[htb!]
\centering
\begin{minipage}{\linewidth}
\begin{lstlisting}[language=JavaScript, basicstyle=\ttfamily\scriptsize, breaklines=true]
// Task: Starting from an empty Scratch project with a single default sprite named 'Sprite1', you should complete the following: 1) When the green flag is clicked, use 'ask and wait' to ask 'What is your name?'. 2) After the user enters an answer, make Sprite1 say exactly the same text they typed for 2 seconds.

return (async function(vm) {
    const timeoutMs = 25000;
    const spriteName = 'Sprite1';
    const testAnswer = 'TestUser123';
    
    // Tracking partial progress
    const results = {
        detectsQuestion: false,
        providesAnswer: false,
        detectsEcho: false,
    };

    let timeoutTimer = null;
    
    try {
        await new Promise((resolve, reject) => {
            // Set up timeout guard
            timeoutTimer = setTimeout(() => reject('Timeout'), timeoutMs);

            // Listen for 'SAY' events from the VM runtime
            vm.runtime.on('SAY', (target, type, text) => {
                if (!target || !target.sprite || target.sprite.name !== spriteName) return;

                const textStr = String(text || '').toLowerCase();
                
                // Step 1: Detect if the sprite asks "name"
                if (!results.detectsQuestion && textStr.includes('name') && textStr.includes('?')) {
                    results.detectsQuestion = true;
                    
                    // Simulate user typing an answer after a short delay
                    setTimeout(() => {
                        vm.runtime.emit('ANSWER', testAnswer);
                        results.providesAnswer = true;
                    }, 500);
                    return;
                }

                // Step 2: Detect if the sprite says the answer back
                if (results.providesAnswer && String(text) === testAnswer) {
                    results.detectsEcho = true;
                    resolve(); // Test Passed!
                }
            });

            // Start the test
            vm.greenFlag();
        });

        // Calculate Score
        const passedTests = Object.values(results).filter(Boolean).length;
        const totalTests = Object.keys(results).length;
        
        return {
            success: passedTests === totalTests,
            score: totalTests ? passedTests / totalTests : 0,
            details: results
        };
        
    } catch (err) {
        return { success: false, reason: err.toString() };
    } finally {
        if (timeoutTimer) clearTimeout(timeoutTimer);
    }
})(window.vm);
\end{lstlisting}
\end{minipage}
\caption{An example of the execution-based evaluation script used in \bench for functional verification.}
\label{fig:eval_script_code}
\end{figure}

\section{Prompts for Semi-Automated Generation}
\label{sec:appendix_prompts}

To ensure the diversity and quality of the tasks in \bench, we utilized LLMs to brainstorm task ideas and generate initial evaluation scripts. The raw outputs were refined by human experts to ensure rigor and solvability, particularly preventing hallucinated block types or logic errors.

\subsection{Task Idea Generation: Create}
\label{app:prompt_create}
For \textbf{Create} tasks, which require constructing projects from scratch, we used the prompt in Figure~\ref{lst:prompt_build} to generate distinct task ideas and detailed requirements, aligned with available Scratch blocks.

\begin{figure}[htb!]
\centering
\begin{minipage}{\linewidth}
\begin{lstlisting}[basicstyle=\ttfamily\scriptsize]
### System / Role
This project is a Scratch Benchmark, containing different task types: build, fix, modify, and algorithm.
- **Create tasks**: Involves building a new project from scratch.

Your job is to generate new tasks of this type: **create**.
Create tasks require the agent to build a project based on the instruction from an empty project with only a default sprite: "Sprite1" (the cat).

### Instructions
1. **Avoid Duplication**: Review the provided overview of existing build tasks to ensure novelty.

2. **Feasibility**: Refer to the standard Scratch block documentation to ensure tasks are feasible.
3. **Format**: Generate configuration files for each task including:
   - Task Name
   - Detailed Instruction
   - Difficulty Level estimate

### Examples (Instructions Only)
- "Starting from an empty Scratch project with a single default sprite named 'Sprite1', you should complete the following: 1) When the green flag is clicked, use 'ask and wait' to ask 'What is your name?'. 2) After the user enters an answer, make Sprite1 say exactly the same text they typed for 2 seconds."
- "Starting from a Scratch project with a single default sprite named 'Sprite1' and 3 backdrops, you should complete the following: 1) When the green flag is clicked, make the stage automatically switch to the next backdrop about every 2 seconds. 2) Ensure the backdrop switching loops forever."
- "Starting from an empty Scratch project with a single default sprite named 'balloon', you should complete the following: 1) Create a variable named 'score' and initialize it to 0 when the green flag is clicked. 2) Each time the balloon is clicked, increase 'score' by 1 and immediately move the balloon to a random position on the stage."

Now, generate five new create tasks.
\end{lstlisting}
\end{minipage}
\caption{Prompt template used for generating Create task ideas during the Expansion stage of benchmark construction. }
\label{lst:prompt_build}
\end{figure}

\subsection{Task Idea Generation: Debug and Extend}
\label{app:prompt_debug_extend}
For \textbf{Debug} and \textbf{Extend} tasks, we leveraged existing Scratch projects as seeds. The prompt in Figure~\ref{lst:prompt_fix_modify} guided the LLM to propose bug injection strategies (for Debug) or feature extensions (for Extend) based on serialized project code.

\begin{figure}[htb!]
\centering
\begin{minipage}{\linewidth}
\begin{lstlisting}[basicstyle=\ttfamily\scriptsize]
### System / Role
Your job is to act as an expert task generator.

### Task Description
Analyze the provided project (Project ID: {project_id}).
The project contains:
1. Project Metadata: Fundamental information about the project.
2. Pseudocode: A text-based representation of the block logic.

**Goal:**
1. Offer 5 ideas on how to "destroy" this project to form **Debug tasks** (introduce logical errors, missing blocks, etc.).
2. Offer 5 ideas for modifying the original functional project to create **Extend tasks** (add features, change behaviors).

### Output Requirements
Output a markdown file with the following structure:
- **Idea Name**: A unique, descriptive name.
- **Description**: Detailed explanation of the bug (for Debug) or feature (for Extend).
\end{lstlisting}
\end{minipage}
\caption{Prompt template used for generating Debug and Extend task ideas during the Expansion stage.}
\label{lst:prompt_fix_modify}
\end{figure}

\subsection{Evaluation Script Generation}
\label{app:prompt_eval}
After human annotators finalized the task definitions, we used the prompt in Figure~\ref{lst:prompt_eval} to generate the initial draft of the JavaScript-based evaluation scripts.

\begin{figure}[htb!]
\centering
\begin{minipage}{\linewidth}
\begin{lstlisting}[basicstyle=\ttfamily\scriptsize]
### System / Role
Your job is to generate evaluation functions for Scratch tasks. These functions validate whether a user's submission (a Scratch project) meets the task requirements.

### Task Information
You are generating an evaluation function for the task: `{task_name}`.
Task Configuration: `{task_config_json}`

### Environment & Tools
- The evaluation function should be a JavaScript file running in a browser environment with access to a global `vm` (Virtual Machine) object.
- **Utility Library**: You have access to `EvaluationUtils` validation helpers (e.g., checking sprite properties, monitoring events, simulating user input).

### Instructions
1. **Structure**:
   - The evaluation should consist of multiple **partial tests** (assertions).
   - Use `cleanupAndFinish(true, resolve)` to mark success.
2. **Logic Check**:
   - Ensure the script validates the *functional outcome* (e.g., "Sprite moves to X=100") rather than just static block presence.

### Example Logic (Say Listener)
If you need to listen for a sprite saying something:
```js
vm.runtime.on('SAY', (target, type, text) => {
    if (target.sprite.name === 'Sprite1' && text.includes('Correct')) {
        console.log('Passed check!');
        cleanupAndFinish(true, resolve);
    }
});
```
\end{lstlisting}
\end{minipage}
\caption{Prompt template used for generating JavaScript evaluation scripts during the Expansion stage.}
\label{lst:prompt_eval}
\end{figure}

\section{System Prompts for Two Interaction Modes}
\label{app:system_prompts}

This section provides the complete system prompts used for both interaction modes in \bench. These prompts define the action spaces, response formats, and execution rules that guide agent behavior.

\subsection{Primitive Mode System Prompt}
\label{app:system_prompt_primitive}

The following system prompt (Figure~\ref{fig:system_prompt_primitive}) is used for primitive mode, where agents interact with the Scratch GUI through low-level UI actions such as click, drag-and-drop, and type.

\begin{figure}[htb!]
\centering
\begin{minipage}{\linewidth}
\begin{lstlisting}[basicstyle=\ttfamily\scriptsize, breaklines=true]
You are an AI assistant that controls the Scratch programming environment using low-level UI actions. Your task is: {INSTRUCTION}

## Low-Level Actions Catalog
Actions (api + args):
- click: args { x?: number, y?: number, index?: number, button?: 'left'|'right'|'middle' }
- double_click: args { x?: number, y?: number, index?: number }
- drag_and_drop: args { start_x?: number, start_y?: number, end_x?: number, end_y?: number, start_index?: number, end_index?: number, duration?: number }
- type: args { text: string }
- key: args { key: string }
- hotkey: args { keys: string[] }
- scroll: args { direction: 'up'|'down'|'left'|'right', amount?: number, x?: number, y?: number, index?: number }
- done: args { }
- failed: args { }

Execution Rules:
1. Targeting: For location-based actions, use EXACTLY ONE method:
   - Preferred: "index" if the UI element is numbered or tagged.
   - Fallback: "x", "y" coordinates for direct screen interaction.
2. Conciseness: Omit any optional parameters that remain at their default values.

## Response Structure Rules

Every response must follow this two-part structure:

1. **Reasoning Section**: Start your response with "Analysis:". Provide a detailed step-by-step reasoning about the current program state and your next move.
2. **Action Section**: Provide exactly ONE fenced code block labeled `json` that contains a single API call JSON only.

Rules:
- The Action Section must be the ONLY code block in your response.
- Do not contain any explanations, comments or trailing commas inside the code block.
- You can only call ONE API per response.

Response format:
Analysis: <Your thoughts here>
```json
{"api": "...", "args": {...}}
```

Response examples:

Example 1:
Analysis: I should click the green flag button by index.
```json
{"api":"click","args":{"index":3}}
```

Example 2:
Analysis: I should type the sprite name into the rename input.
```json
{"api":"type","args":{"text":"Sprite1"}}
```

Example 3:
Analysis: I have finished the task.
```json
{"api":"done"}
```

## Per-Turn Input

Each turn, you will receive a screenshot and the element information of the current UI.

The element information is provided as a unified list of all visible UI elements, including:
- index: Sequential 0-based identifier for each UI element (use this for index-based actions)
- type: Element category (canvas, inputs, sprites, blocks, green_flag, stop_button, stage, etc.)
- text: Visible text content or element description (may include values, placeholders, labels)
- position: Location and size as (x, y) widthxheight in pixels
\end{lstlisting}
\end{minipage}
\caption{Complete system prompt for primitive mode. This prompt defines the low-level action space (click, drag-and-drop, type, etc.) and the structured response format required for GUI-based interaction.}
\label{fig:system_prompt_primitive}
\end{figure}

\clearpage

\subsection{Composite Mode System Prompt}
\label{app:system_prompt_composite}

The following system prompt (Figures~\ref{fig:system_prompt_composite_part1} and \ref{fig:system_prompt_composite_part2}) is used for composite mode, where agents interact with Scratch through high-level semantic APIs that abstract away GUI operations. Due to length, the prompt is shown in two parts.

\begin{figure}[htb!]
\centering
\begin{minipage}{\linewidth}
\begin{lstlisting}[basicstyle=\ttfamily\tiny, breaklines=true]
You are an AI assistant that controls the Scratch programming environment using high-level APIs. Your task is: {INSTRUCTION}

## Encapsulated API Catalog
APIs (api + args):
- select_sprite: args { name: string }
  Switch the editing target to the named sprite.
- select_stage: args { }
  Switch the editing target to the Stage.
- add_variable: args { name: string, scope: 'sprite'|'all' }
  Create a variable scoped to the current sprite or global.
- add_list: args { name: string, scope: 'sprite'|'all' }
  Create a list scoped to the current sprite or global.
- add_block: args { blockType: string, creation?: { variableName?: string, listName?: string } }
  Create a new block in the current workspace, with optional variable/list binding.
- connect_blocks: args { sourceBlockIndex: number, targetBlockIndex: number, placement: { kind: 'stack_before'|'stack_after'|'statement_into'|'value_into'|'wrap', inputName?: string } }
  Connect one block to another (stack, insert into C-slot, value slot, or wrap).
- detach_blocks: args { blockIndex: number }
  Detach a block (and its stack) into a new top-level stack.
- set_block_field: args { blockIndex: number, fieldName: string, value: any }
  Set a block field (e.g., dropdown, text, number) to a new value.
- delete_block: args { blockIndex: number }
  Delete a block (and its attached stack, if any).
- done: args { }
  Mark task as completed.
- failed: args { }
  Mark task as failed.

Execution Rules:
1. add_block: args.creation.variableName is required when blockType='data_variable'.
2. add_block: args.creation.listName is required when blockType='data_listcontents'.
   For other blockType values, creation.variableName and creation.listName are not supported and are ignored.
3. connect_blocks: placement.inputName is required when kind is 'statement_into' or 'value_into'.
4. connect_blocks: value slots (e.g., CONDITIONS, numeric/string inputs) must use kind='value_into'; statement slots (C-shaped SUBSTACK/SUBSTACK2) use kind='statement_into'.
5. detach_blocks: detaches the specified block and its following stack into a new top-level stack.

## Response Structure Rules

Every response must follow this two-part structure:

1. **Reasoning Section**: Start your response with "Analysis:". Provide a detailed step-by-step reasoning about the current program state and your next move.
2. **Action Section**: Provide exactly ONE fenced code block labeled `json` that contains a single API call JSON only.
Rules:
- The Action Section must be the ONLY code block in your response.
- Do not contain any explanations, comments or trailing commas inside the code block.
- You can only call ONE API per response.

Response format:
Analysis: <Your thoughts here>
```json
{"api": "...", "args": {...}}
```

Response examples:

Example 1:
Analysis: I will connect block 2 to block 1 as the next stack item.
```json
{"api":"connect_blocks","args":{"sourceBlockIndex":2,"targetBlockIndex":1,"placement":{"kind":"stack_after"}}}
```

Example 2:
Analysis: I will switch to the Stage as the current target.
```json
{"api":"select_stage"}
```

Example 3:
Analysis: I will update a field on block index 3.
```json
{"api":"set_block_field","args":{"blockIndex":3,"fieldName":"QUESTION","value":"Input:"}}
```

(continued in Figure~\ref{fig:system_prompt_composite_part2})
\end{lstlisting}
\end{minipage}
\caption{Composite mode system prompt (Part 1/2): API Catalog and Response Structure Rules. This part defines the high-level API action space including block manipulation commands (add block, connect blocks, set field, delete block, etc.) and the structured response format.}
\label{fig:system_prompt_composite_part1}
\end{figure}

\begin{figure}[htb!]
\centering
\begin{minipage}{\linewidth}
\begin{lstlisting}[basicstyle=\ttfamily\tiny, breaklines=true]
(continued from Figure~\ref{fig:system_prompt_composite_part1})

## Scratch Blocks Catalog

All available blocks:
Control: control_start_as_clone, control_repeat, control_repeat_until, control_forever, control_wait, control_wait_until, control_if, control_if_else, control_stop, control_create_clone_of, control_delete_this_clone
Variables: data_variable, data_setvariableto, data_changevariableby, data_showvariable, data_hidevariable, data_listcontents, data_addtolist, data_deleteoflist, data_deletealloflist, data_insertatlist, data_replaceitemoflist, data_itemoflist, data_itemnumoflist, data_lengthoflist, data_listcontainsitem, data_showlist, data_hidelist
Events: event_whenflagclicked, event_whenkeypressed, event_whenthisspriteclicked, event_whenbackdropswitchesto, event_whengreaterthan, event_whenbroadcastreceived, event_broadcast, event_broadcastandwait
Looks: looks_say, looks_sayforsecs, looks_think, looks_thinkforsecs, looks_show, looks_hide, looks_switchcostumeto, looks_nextcostume, looks_switchbackdropto, looks_nextbackdrop, looks_changeeffectby, looks_seteffectto, looks_cleargraphiceffects, looks_changesizeby, looks_setsizeto, looks_gotofrontback, looks_goforwardbackwardlayers, looks_size, looks_costumenumbername, looks_backdropnumbername
Motion: motion_movesteps, motion_gotoxy, motion_goto, motion_turnright, motion_turnleft, motion_pointindirection, motion_pointtowards, motion_glidesecstoxy, motion_glideto, motion_ifonedgebounce, motion_setrotationstyle, motion_changexby, motion_setx, motion_changeyby, motion_sety, motion_xposition, motion_yposition, motion_direction
Operators: operator_add, operator_subtract, operator_multiply, operator_divide, operator_random, operator_mod, operator_round, operator_mathop, operator_join, operator_letter_of, operator_length, operator_lt, operator_equals, operator_gt, operator_and, operator_or, operator_not, operator_contains
Sensing: sensing_resettimer, sensing_setdragmode, sensing_askandwait, sensing_timer, sensing_mousex, sensing_mousey, sensing_dayssince2000, sensing_current, sensing_mousedown, sensing_keypressed, sensing_touchingobject, sensing_touchingcolor, sensing_coloristouchingcolor, sensing_distanceto, sensing_of, sensing_loudness, sensing_answer, sensing_username
Sound: sound_play, sound_playuntildone, sound_stopallsounds, sound_seteffectto, sound_changeeffectby, sound_cleareffects, sound_setvolumeto, sound_changevolumeby, sound_volume

## Per-Turn Input

Each turn, you will receive information about the current editing target and its blocks pseudocode.

Format notes of blocks pseudocode:
- [top] means the TOP of a block stack. The following blocks (until a blank line or next [top]) are stacked under it in order.
- Each block line starts with `#<index> <opcode> [field=value ...]`. The index is a local identifier for reference only.
- Fields of a block are shown as `- <NAME>: <value>` followed by indented lines showing the nested reporter/boolean blocks.
- Indentation indicates nesting hierarchy. Increased indentation denotes content nested within that input or substack.
- Some fields may show `<choices: ...>` to list available values.
\end{lstlisting}
\end{minipage}
\caption{Composite mode system prompt (Part 2/2): Scratch Blocks Catalog and Input Format. This part provides the complete catalog of available Scratch blocks organized by category (Control, Variables, Events, Looks, Motion, Operators, Sensing, Sound) and explains the pseudocode observation format.}
\label{fig:system_prompt_composite_part2}
\end{figure}

\section{Element List Construction and OCR--DOM Fusion}
\label{app:element-list-fusion}

\paragraph{DOM element extraction.}
We query a fixed set of CSS selectors via a batch API and return a flat list of elements. 
Each element includes a type (the selector name), text, and bounding box (x, y, width, height).
Elements outside the viewport or visually occluded are filtered out. 
Text is collected by prioritizing aria-label, title, and visible text; for Scratch blocks, we further resolve block identifiers when available.

\paragraph{OCR extraction.}
Given the screenshot, we call the OCR service (PaddleOCR~\citep{cui2025paddleocr}) to detect text regions and their bounding boxes. 
OCR elements below a confidence threshold are discarded.

\paragraph{Fusion and deduplication.}
We normalize DOM and OCR elements to a common structure and merge them.
Deduplication is performed geometrically: OCR boxes fully covered by a DOM box are removed (with a special case for blocks on the canvas).
The final list preserves DOM order and appends the remaining OCR-only text elements.

\section{Use of LLMs}\label{sec:llm}
We use LLMs for polish writing. 
Specifically, LLMs assist in refining the grammar, clarity, and overall presentation of the paper, ensuring that the text is clear and professionally written. 
No experimental results or core content were generated by LLMs.

\end{document}